\documentclass[nonblindrev]{informs3} 

\usepackage{endnotes}
\let\footnote=\endnote

\usepackage{natbib}
 \bibpunct[, ]{(}{)}{,}{a}{}{,}%
 \def\bibfont{\small}%
\TheoremsNumberedThrough     
\ECRepeatTheorems

\EquationsNumberedThrough    

\usepackage{graphicx}
\usepackage{amssymb}

\usepackage[USenglish]{babel}

\usepackage{booktabs} 

\usepackage[title]{appendix}
\usepackage[T1]{fontenc}
\usepackage{times}
\usepackage{graphicx}
\usepackage{amsmath}
\usepackage{amsfonts}
\usepackage{wasysym}
\usepackage{url}
\usepackage{makecell}
\usepackage{rotating}
\usepackage[boxed,vlined,linesnumbered,ruled]{algorithm2e}
\let\oldnl\nl
\newcommand{\nonl}{\renewcommand{\nl}{\let\nl\oldnl}}
\usepackage{color}
\usepackage{ucs}
\usepackage{subfigure}
\usepackage{latexsym}
\usepackage{multirow}
\usepackage[autolanguage]{numprint}
\usepackage{array}
\usepackage{times}
\usepackage{xcolor,colortbl}
\definecolor{green}{rgb}{0.1,0.1,0.1}

\usepackage[utf8x]{inputenc}
\definecolor{darkblue}{HTML}{000099}
\usepackage[colorlinks=true,linkcolor=black,citecolor=darkblue]{hyperref}
\DeclareMathOperator{\fst}{fst}
\DeclareMathOperator{\prv}{prv}
\DeclareMathOperator{\nxt}{nxt}
\DeclareMathOperator{\w}{w}
\DeclareMathOperator{\s}{s}
\DeclareMathOperator{\departureTime}{departureTime}
\DeclareMathOperator{\discardedTime}{discardedTime}
\DeclareMathOperator{\load}{load}
\usepackage{enumerate}

\begin{document}


\RUNAUTHOR{Saint-Guillain, Solnon, and Deville}

\RUNTITLE{PFS for the SS-VRPTW with both Random Customers and Reveal Times}

\TITLE{Progressive Focus Search for the Static and Stochastic VRPTW with both Random Customers and Reveal Times}

\ARTICLEAUTHORS{%
\AUTHOR{Michael Saint-Guillain}
\AFF{Université catholique de Louvain, Belgium, \EMAIL{michael.saint@uclouvain.be}}
\AUTHOR{Christine Solnon}
\AFF{Institut National des Sciences Appliquées de Lyon, France, \EMAIL{christine.solnon@insa-lyon.fr}}
\AUTHOR{Yves Deville}
\AFF{Université catholique de Louvain, Belgium}
} 


\ABSTRACT{
Static stochastic VRPs aim at modeling real-life VRPs by considering uncertainty on data. In particular, the SS-VRPTW-CR considers stochastic customers with time windows and does not make any assumption on their reveal times, which are stochastic as well. Based on customer request probabilities, we look for an a priori solution composed preventive vehicle routes, minimizing the expected number of unsatisfied customer requests at the end of the day. A route describes a sequence of strategic vehicle relocations, from which nearby requests can be rapidly reached. Instead of reoptimizing online, a so-called recourse strategy defines the way the requests are handled, whenever they appear. 
In this paper, we describe a new recourse strategy for the SS-VRPTW-CR, improving vehicle routes by skipping useless parts. We show how to compute the expected cost of a priori solutions, in pseudo-polynomial time, for this recourse strategy. We introduce a new meta-heuristic, called Progressive Focus Search (PFS), which may be combined with any local-search based algorithm for solving static stochastic optimization problems. PFS accelerates the search by using approximation factors: from an initial rough simplified problem, the search progressively focuses to the actual problem description. We evaluate our contributions on a new, real-world based, public benchmark.
}

\maketitle

\section*{Introduction \label{sec:Introduction}}
In the Vehicle Routing Problem with Time Windows, a set of customers must be serviced by a homogeneous fleet of capacitated vehicles, while reconciling each customer's time windows and vehicle travel times, as well as cumulated customers' demands and vehicle capacities. 
Whereas deterministic VRP(TW)s assume perfect information on input data, in real-world applications some input data may be uncertain when computing a solution. 
In this paper, we focus on cases where the customer presence is a priori unknown. Furthermore, we assume to be provided with some probabilistic knowledge on the missing data. 
In many situations, the probability distributions can be obtained from historical data.
In order to handle new customers who appear dynamically, the current solution must be adapted as such random events occur. 
Depending on the operational context, we distinguish two fundamentally different assumptions.
If the currently unexecuted part of the solution can be arbitrarily redesigned, then we are facing a {\em Dynamic and Stochastic VRP(TW)} (DS-VRP(TW)). 
In that case, the solution is adapted by re-optimizing the new current problem while fixing the executed partial routes.

\begin{figure}
\centering
\includegraphics[width=0.8\textwidth]{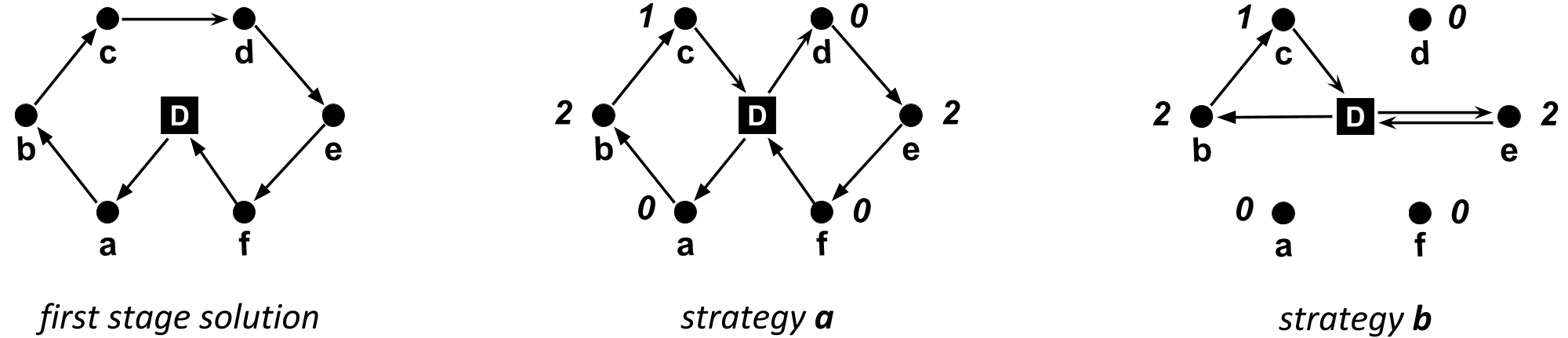}
\caption{
Recourse strategies for the SS-VRP with stochastic customers and demands (\citeauthor{Bertsimas1992}, \citeyear{Bertsimas1992}). 
The vehicle has a capacity of 3. 
The first stage solution states the {\em a priori} sequence of customer visits.
When applying strategy {\bf \textit{a}}, the vehicle unloads at the depot after visiting {\bf c}.
In strategy {\bf \textit{b}}, absent customers ({\bf a}, {\bf d}, {\bf f}) are skipped.
}
\label{fig:example_bertsimas}
\end{figure}

If the routes can only be adapted by following some predefined scheme, then we are facing a {\em Static and Stochastic VRP(TW)} (SS-VRP(TW)).
In the SS-VRP(TW), whenever a bit of information is revealed, the current solution is adapted by applying a \emph{recourse strategy}.
Based on the probabilistic information, we seek a first stage (also called \emph{a priori}) solution that minimizes its a priori cost, plus the expected sum of penalties caused by the recourse strategy. 
In order for the evaluation function to remain tractable, the recourse strategy must be efficiently computable, hence simple enough to avoid re-optimization.
For example, in \cite{Bertsimas1992} the customers are known, whereas their demands are revealed online. 
Two different assumptions are considered, leading to different recourse strategies, as illustrated in Fig. \ref{fig:example_bertsimas}. 
In strategy \textbf{a}, each demand is assumed to be revealed when the vehicle arrives at the customer place. 
If the vehicle reaches its maximal capacity, then the first stage solution is adapted by adding a round trip to the depot. 
In strategy \textbf{b}, each demand is revealed when leaving the previous customer, allowing to skip customers having null demands.

In the recent review of \cite{Gendreau2016}, the authors argue for new recourse strategies: With the increasing use of ICT, customer information is likely to be revealed on a very frequent basis. 
In this context, the chronological order in which this information is transmitted no longer matches the planned sequences of customers on the vehicle routes.
In particular, the authors consider as paradoxical the fact that the existing literature on SS-VRPs with random Customers (SS-VRP-C) assumes full knowledge on the presence of customers at the beginning of the operational period.

In this paper, we focus on the SS-VRPTW with both random Customers and Reveal times (SS-VRPTW-CR) introduced by \cite{Saint-Guillain2017}, in which no assumption is made on the moment at which a request is known.
The goal is to compute the first-stage solution that minimizes the expected number of rejected requests, while avoiding assumptions on the moment at which customer requests are revealed.
To handle uncertainty on the reveal times, waiting (re)locations are part of first-stage solutions.

\subsubsection*{Application example.}
Let us consider the problem of managing a team of on-duty doctors, operating at patient home places during nights and week-ends.
On-call periods start with all the doctors at a central depot, where each get assigned a taxi cab for efficiency and safety. 
Patient requests arrive dynamically. We know from historical data the probability that a request appears depending on the region and time of day.
Each online request comes with a hard deadline, and a recourse strategy can be used to decide whether the request can be satisfied in time and how to adapt the routes accordingly.
If it cannot be handled in time, the request is rejected and entrusted to an (expensive) external service provider. 
The goal is to minimize the expected number of rejected requests.
In such context, involving very short deadlines, relocating idle doctors anticipatively is often critical.
Modeled as a SS-VRPTW-CR, it is possible to compute a first-stage solution composed of (sequences of) waiting locations, optimizing the expected quality of service.

\subsubsection*{Contributions.}
We introduce an improved recourse strategy for the SS-VRPTW-CR that optimizes routes by skipping some useless parts. 
Closed-form expressions are provided to efficiently compute expected costs for the new recourse strategy.
Another contribution is a new meta-heuristic, called {\em Progressive Focus Search} (PFS), for solving static stochastic optimization problems. 
PFS accelerates the solution process by using approximation factors, both reducing the size of the search space and the complexity of the objective function. 
These factors are progressively decreased during the search process so that, from an initial rough approximation of the problem, the search gradually focuses to the actual problem. 
We also introduce a new public benchmark for the SS-VRPTW-CR, based on real-word data coming from the city of Lyon. 
Experimental results on this benchmark show that PFS obtains better results than a classical search. 
Important insights are brought to light.
By comparing with a basic (yet realistic) policy which does not exploit stochastic knowledge, we show that our stochastic models are particularly beneficial when the number of vehicles increases and when time windows are tight. 

\subsubsection*{Organization.}
In Section \ref{sec:related-work}, we review existing studies on VRPs with stochastic customers and clearly position the SS-VRPTW-CR with respect to them.
In Section \ref{sec:Problem-description}, we formally define the general SS-VRPTW-CR. 
Section \ref{sec:recourse} presents a new recourse strategy, which we present as both a generalization and an improvement over the recourse strategy previously proposed in \cite{Saint-Guillain2017}. 
In Section \ref{sec:PFS}, we describe the {\em Progressive Focus Search} metaheuristic for static stochastic optimization problems and show how to instantiate it to solve the SS-VRPTW-CR.
In Section \ref{sec:bench_experimental_plan}, we describe a new public benchmark for the SS-VRPTW-CR, derived from real-world data, and describe the experimental settings.
The experimental results are analyzed in Section \ref{sec:small_instances} for small instances and in Section \ref{sec:big_instances} for larger instances. 
Finally, further research directions are discussed in Section \ref{sec:conclusions}.

\section{Related work \label{sec:related-work}}

By definition, the SS-VRPTW-CR is a static problem. Decisions are made \emph{a priori}. 
The reader interested in \emph{online} decision making should refer to Dynamic and Stochastic VRPs, such as the DS-VRPTW. 
A recent literature review about DS-VRPs can be found in \cite{ritzinger2015survey}.
\cite{Gendreau2016} provides a literature review on Static and Stochastic VRPs (SS-VRP). 
According to \cite{pillac2013review}, the most studied cases in SS-VRPs are: 
\begin{itemize}

\item Stochastic times (SS-VRP-T), where travel and/or service times are random; see \textit{e.g.} 
\cite{kenyon2003stochastic}, 
\cite{verweij2003sample}, 
\cite{li2010vehicle}. 

\item Stochastic demands (SS-VRP-D), where all customers are known in advance but their demands are random variables; see \textit{e.g.} 
\cite{Laporte2002}, 
\citeauthor{Mendoza2010} (\citeyear{Mendoza2010}, \citeyear{Mendoza2011}),  
\citeauthor{Secomandi2000} (\citeyear{Secomandi2000}, \citeyear{Secomandi2009}) 
and \cite{Gauvin2014}.

\item Stochastic customers (SS-VRP-C), where customer presences are described by random variables. Since the SS-VRPTW-CR belongs to this category, this non-exhaustive literature review is limited to  to studies involving customer uncertainty.
\end{itemize}



The Traveling Salesman Problem (TSP) is a special case of the VRP with only one uncapacitated vehicle.
\cite{jaillet1985probabilistic} formally introduced the TSP with stochastic Customers (SS-TSP-C), a.k.a. the probabilistic TSP (PTSP) or TSPSC in the literature, and provided mathematical formulations and a number of properties and bounds of the problem (see also \citeauthor{Jaillet1988}, \citeyear{Jaillet1988}). 
In particular, he showed that an optimal solution for the deterministic problem may be arbitrarily bad in case of uncertainty. 
\cite{laporte1994priori} developed the first exact solution method for the SS-TSP-C.
Heuristics for the SS-TSP-C have then been proposed in 
\cite{jezequel1985probabilistic}, 
\cite{rossi1987aspects}, 
\cite{bertsimas1995approaches},  
and \cite{Bianchi2007} 
as well as meta-heuristics, such as simulated annealing (\cite{Bowler2003}) or ant colony optimization (\citeauthor{bianchi2002aco}, \citeyear{bianchi2002aco}). 

Particularly close to the SS-VRPTW-CR is the SS-TSP-C with Deadlines introduced by \cite{Campbell2008x}. 
Unlike the SS-VRPTW-CR, authors assume that customer presences 
are revealed all at once at the beginning of the day.
They showed that deadlines are particularly challenging when considered in a stochastic context, and proposed two recourse strategies to address deadline violations. 
A survey on the SS-TSP-C may be found in \cite{Henchiri2014}. 

The first SS-VRP-C has been studied by \cite{jezequel1985probabilistic} 
as a generalization of the SS-TSP-C.
\cite{Bertsimas1992} considered a VRP with stochastic Customers and Demands (SS-VRP-CD), as described in the introduction section. 
\cite{gendreau1995exact} developed the first exact algorithm for solving the SS-VRP-CD for instances up to 70 customers, by means of an integer L-shaped method, and 
\cite{gendreau1996tabu} later proposed a tabu search algorithm.
A preventive restocking strategy for the SS-VRP with random demands has been proposed by \cite{Yang2000}. 
\cite{Biesinger2016} later introduced a variant for the Generalized SS-VRP with random demands.

\cite{Sungur2010} considered the Courier Delivery Problem with Uncertainty.
Potential customers have deterministic soft time windows but are present probabilistically, with uncertain service times. 
The objective is to construct an {\em a priori} solution, to be used every day as a basis, then adapted to daily requests.
Unlike the SS-VRPTW-CR, the set of customers is revealed at the beginning of the operations.   
\cite{heilporn2011integer} introduced the Dial-a-Ride Problem (DARP) with stochastic customer delays. 
Each customer is present at its pickup location with a stochastic delay. 
A customer is then skipped if it is absent when the vehicle visits the corresponding location, involving the cost of fulfilling the request by an alternative service (\emph{e.g.}, a taxi). 
In a sense, stochastic delays imply that each request is revealed at some uncertain time during the planning horizon. 
That study is thus related to our problem, except that in the SS-VRPTW-CR, part of the requests will reveal to never appear. 


\section{Problem description: the SS-VRPTW-CR\label{sec:Problem-description}}

This section recalls the definition of the SS-VRPTW-CR, initially provided in \cite{Saint-Guillain2017}.
In fact, it contains parts taken from section 3 of the aforementioned paper. 

\subsubsection*{Input data.}

We consider a complete directed graph $G=(V,A)$ and a discrete time horizon $H=[1,h]$, where the interval $[a,b]$ denotes the set of all integer values $i$ such that $a \le i \le b$. 
A travel time (or distance) $d_{i,j} \in \mathbb{N}$ is associated with every arc $(i,j) \in A$. 
The set of vertices $V=\{0\} \cup W \cup C$ is composed of a depot $0$, a set of $m$ waiting locations $W=[1,m]$, and a set of $n$ customer vertices $C=[m+1,m+n]$.
We note $W_0=W \cup \{0\}$ and $C_0=C \cup \{0\}$. 
The fleet is composed of $K$ vehicles of maximum capacity $Q$. 
Let $R=C \times H$ be the set of potential requests. An element $r = (c,\Gamma)$ of $R$ represents a potential request revealed at time $\Gamma$ at customer vertex $c$.
It is associated the following deterministic attributes: a demand $q_r\in [1,Q]$, a service duration $s_r \in H$ and a time window $[e_r,l_r]$ with $\Gamma \le e_r \le l_r \le h$. 
We note $p_r$ the probability that $r$ appears on vertex $c$ at time $\Gamma$ and assume independence between requests.
Although our formalism imposes $\Gamma \ge 1$ for all potential requests, in practice a request may be known with probability $1$, leading to a deterministic request. 
Finally, different customers in $C$ can share the same geographical location, making it possible to consider different types of requests in terms of deterministic attributes.
To simplify notations, we use $\Gamma_r$ to denote the reveal time of a request $r \in R$ and $c_r$ for its customer vertex.
Furthermore, a request $r$ may be written in place of its own vertex $c_r$. 
For instance, the distance $d_{v,c_r}$ may also be written as $d_{v,r}$.
Table \ref{table:notations} summarizes the main notations.
\begin{table}[h]
\caption{\label{table:notations}
Notation summary: graph and potential requests.}
\begin{tabular}{lllll}
\hline
$G=(V,A)$ 		& Complete directed graph						&~~~~& $R=C \times H$ 	& Set of potential requests	\\
$V=\{0\} \cup W \cup C$	& Set of vertices (depot is $0$)		&& $\Gamma_r$			& Reveal time of request $r \in R$	\\
$W=[1,m]$		& Waiting vertices 								&& $c_r$						& Customer vertex hosting request $r \in R$	\\
$C=[m+1,m+n]$	& Customer vertices 							&& $s_r$						& Service time of request $r \in R$	\\	
$d_{i,j}$ 		& Travel time of arc $(i,j) \in A$				&& $[e_r,l_r]$					& Time window of request $r \in R$	\\
$K$				& Number of vehicles 		 					&& $q_r$						& Demand of request $r\in R$\\
$Q$				& Vehicle capacity								&& $p_r$ 						& Probability associated with request $r$\\
$H = [1,h]$		& Discrete time horizon		 					\\
\hline
\end{tabular}
\end{table}

\subsubsection*{First-stage solution.}

The first-stage solution is computed offline, before the beginning of the time horizon.
It consists of a set of $K$ vehicle routes visiting a subset of the $m$ waiting vertices, together with duration variables denoted by $\tau$ indicating how long a vehicle should wait on each vertex. 
More specifically, we denote by $(x,\tau)$ a first-stage solution to the SS-VRPTW-CR.
$x = \{ x_1, ..., x_K \}$ defines a set of $K$ disjoint sequences of waiting vertices of $W$, each starting and ending with the depot. 
Each vertex of $W$ occurs at most once in $x$.
We note $W^x \subseteq W$, the set of waiting vertices visited in $x$. 
The vector $\tau$ associate a waiting time $\tau_w \in H$ with every waiting vertex $w \in W^x$.
For each sequence $x_k = \langle w_{m_1}, ..., w_{m_k} \rangle$, the vehicle is back at the depot by the end of the time horizon: 
$$\sum_{i=1}^{k-1} d_{w_{m_i},w_{m_{i+1}}} + \sum_{i=2}^{k-1} \tau_{w_{m_i}} \le h $$
In other words, $x$ defines a solution to a \emph{Team Orienteering Problem} (TOP, see \cite{Chao1996}) to which each visited location is assigned a waiting time by $\tau$.
Given a first-stage solution $(x,\tau)$, we define $on(w) = [\underline{on}(w), \overline{on}(w)]$ for each vertex $w \in W^x$ such that $\underline{on}(w)$ (resp. $\overline{on}(w)$) is the arrival (resp. departure) time at $w$.
In a sequence $\langle w_{m_1}, ..., w_{m_k}\rangle$ in $x$, we then have
$\underline{on}(w_{m_i}) = \overline{on}(w_{m_{i-1}}) + d_{w_{m_{i-1}},w_{m_i}}$ and $\overline{on}(w_{m_i}) = \underline{on}(w_{m_i}) + \tau_{w_{m_i}} $
for $i \in [2,k]$ and assume that $\overline{on}(w_{m_1}) = 1$.

\subsubsection*{Recourse strategy and optimal first-stage solution}
Given a first stage solution $(x, \tau)$, a recourse strategy states how the requests, which appear dynamically,  are handled by the vehicles.
In other words, it defines how the second-stage solution is gradually constructed, based on $(x, \tau)$ and depending on these online requests.
A more formal description of recourse strategies in the context of the SS-VRPTW-CR is provided in \cite{Saint-Guillain2017}.
Let a recourse strategy $\mathcal{R}$. 
An optimal first-stage solution $(x,\tau)$ to the SS-VRPTW-CR minimizes the expected cost of the second-stage solution:
\begin{align}
\text{(SS-VRPTW-CR)}\hspace{1em} & \underset{x,\tau}{\text{Minimize}} \hspace{0.5em} \mathcal{Q^R}(x,\tau) \hspace{1em} \label{eq:ss-vrptw-general} \\
& \text{s.t.} \hspace{1em} ~(x,\tau) \text{ is a first-stage solution}. \label{eq:ss-vrptw-general-2} 
\end{align}
The objective function $\mathcal{Q^R}(x,\tau)$, which is nonlinear in general, is the \emph{expected number of rejected requests}, {\em i.e.}, requests that fail to be visited under recourse strategy $\mathcal{R}$ and first-stage solution $(x,\tau)$.
Note that $\mathcal{Q^R}(x,\tau)$ actually represents an expected quality of service, which does not take travel costs into account. 
In fact, in most practical applications that could be formulated as an SS-VRPTW-CR, quality of service prevails whenever the number of vehicles is fixed, as travel costs are usually negligible compared to the labor cost of the mobilized mobile units. 

Formulation (\ref{eq:ss-vrptw-general})-(\ref{eq:ss-vrptw-general-2}) states the problem in general terms, hiding two non-trivial issues. 
Given a recourse strategy $\mathcal{R}$, finding a computationally tractable way to evaluate $\mathcal{Q^R}$ constitutes the first challenge. We address it in Section \ref{sec:recourse}, based on a new recourse strategy we propose. 
The second problem naturally concerns the minimization problem, or how to deal with the solution space. This is addressed in Section \ref{sec:PFS}. 
For completeness, a mathematical formulation of the constraints involved by (\ref{eq:ss-vrptw-general-2}) is provided in Appendix \ref{sec:sip}.

\section{A new recourse strategy \label{sec:recourse}}

The strategy we introduce, called $\mathcal{R}^{q+}$, is a generalization and an improvement of strategy $\mathcal{R}^{\infty}$ introduced in \cite{Saint-Guillain2017}. 
First, it generalizes $\mathcal{R}^{\infty}$ by taking vehicle capacities into account. 
Second, $\mathcal{R}^{q+}$ improves $\mathcal{R}^{\infty}$ by saving operational time when possible, by avoiding some pointless round trips from waiting vertices.

For the sake of completeness and since $\mathcal{R}^{q+}$ generalizes $\mathcal{R}^{\infty}$, part of this section includes elements from section 4 of \cite{Saint-Guillain2017}, which are common to both strategies.
We emphasize the common points and differences between these two strategies at the end of this section.

\subsection{Description of $\mathcal{R}^{q+}$}
Informally, the recourse strategy $\mathcal{R}^{q+}$ accepts a request revealed at time $t$ if the assigned vehicle is able to adapt its first-stage tour to visit the customer, given its set of previously accepted requests. 
Time window and capacity constraints should be respected, and already accepted requests should not be disturbed.

Ideally, whenever a request appears and prior to determine whether it can be accepted, a vehicle should be selected to minimize objective function (\ref{eq:ss-vrptw-general}). 
Furthermore, if several requests appear at the same time unit and amongst the subset of these that are possibly acceptable, some may not contribute optimally to (\ref{eq:ss-vrptw-general}). 
Given a set of accepted requests, the order in which they are handled also plays a critical role. 
Unfortunately, none of these decisions can be made optimally without reducing to a NP-hard problem.
In order for $\mathcal{R}^{q+}$ to remains efficiently computable, they are necessarily made heuristically.


The solution proposed in \cite{Saint-Guillain2017} makes these decisions beforehand. 
Before the start of the operations and in order to avoid reoptimization, the set $R$ of potential requests is ordered. Each potential request $r \in R$ is also preassigned to exactly one planned waiting vertex in $W^x$, and therefore one vehicle, based on geographical considerations.

\subsubsection{Request ordering} The ordering heuristic is independent of the current first-stage solution. Different orders may be considered, provided that the order is total, strict, and $\forall r_1, r_2 \in R$, if the reveal time of $r_1$ is smaller than the reveal time of $r_2$ then $r_1$ must be smaller than $r_2$ in the request order.
We order $R$ by increasing reveal time first, end of time window second, and lexicographic order to break further ties. 

\subsubsection{Request assignment according to a first-stage solution}
Given a first-stage solution $(x,\tau)$, we assign each request of $R$ either to a waiting vertex visited in $x$ or to $\bot$ to denote that $r$ is not assigned. We note $\w: R\rightarrow W^x\cup \{\bot\}$ this assignment.
It is computed for each first-stage solution $(x, \tau)$ before the application of the recourse strategy. 
To compute this assignment, for each request $r$, we first compute the set $W^x_r$ of waiting vertices from which satisfying $r$ is possible, if $r$ appears: 
\[W^x_r = \{ w \in W^x : t^\text{min}_{r,w} \le t^\text{max}_{r,w} \} \notag\]
where $t^\text{min}_{r,w}$ and $t^\text{max}_{r,w}$ are defined as follows.
Time $t^\text{min}_{r, v}= \max\{\underline{on}(\w(r)), ~ \Gamma_{r}, ~ e_{r} - d_{v,{r}}\}$ is the earliest time at which the vehicle can possibly leave vertex $v \in C \cup W$ in order to satisfy request $r$. 
Time $t^\text{max}_{r,v} = \min \{ l_{r} - d_{v,r} , ~ \overline{on}(s(\w(r))) - d_{v,r} - s_{r} - d_{r, s(\w(r))} \}$, where $s(\w(r))$ is the waiting vertex that directly follows $\w(r)$ in the first-stage solution $(x,\tau)$, is the latest time at which a vehicle can leave vertex $v$ to handle $r$
and arrive at $s(\w(r))$ in time.
Given the set $W^x_r$ of feasible waiting vertices for $r$, we define the waiting vertex $\w(r)$ associated with $r$ as follows:
\begin{itemize}
\item If $W^x_r = \emptyset$, then $\w(r)=\bot$ ($r$ is always rejected as it has no feasible waiting vertex);
\item Otherwise, $\w(r)$ is set to the feasible vertex of $W^x_r$ that has the least number of requests already assigned to it (further ties are broken with respect to vertex number). This heuristic rule aims at evenly distributing potential requests on waiting vertices.
\end{itemize}
Once finished, the request assignment ends up with a partition $\{ \pi_\bot, \pi_1, ..., \pi_K \}$ of $R$,
where $\pi_k$ is the set of requests assigned to the waiting vertices visited by vehicle $k$ and $\pi_\bot$ is the set of unassigned requests (such that $\w(r)=\bot$).
We note $\pi_w$, the set of requests assigned to a waiting vertex $w \in W^x$.

\subsubsection{Using $\mathcal{R}^{q+}$ to adapt a first-stage solution at time $t$}

At each time step $t$, the recourse strategy is applied to decide whether to accept or reject the new incoming requests, if any, and determine the appropriate vehicle actions. 
Let $A^{t-1}$ be the set of \emph{accepted} requests up to time $t-1$.
Note that $A^{t-1}$ is likely to contain some requests that have been accepted but are not yet satisfied (\textit{i.e.} not yet visited). 

\paragraph{Availability time.}
The decision to accept or reject a request $r\in\pi^k$ appearing at time $t = \Gamma_r$ depends on when vehicle $k$ will be available for $r$. 
By available, we mean that it has finished serving all its accepted requests that precede $r$, according to the predefined order on $R$. 
This time is denoted by ${\it available}(r)$.
It is only defined when all the accepted requests, that must be served before $r$ by the same vehicle, are known.
If $r$ is the first request of its waiting vertex, the first of $\pi_{\w(r)}$, then: 
$$available(r) = \underline{on}(w), w = \w(r) = \w(r) .$$
Otherwise, let $r^-$ be the request that directly precedes $r$ in $\pi_{\w(r)}$. 
As the requests assigned to $\w(r)$ are ordered by increasing reveal time, we know all these accepted requests for sure when $t\geq \Gamma_{r^-}$. 
Given current time $t \ge \Gamma_r$, function $available(r)$ is defined in $\mathcal{R}^{q+}$ as:
\begin{align}
available(r) 	&= \begin{cases}
		\max \big( available(r^-)+ d_{v(r^-), r^-}, e_{r^-} \big) + s_{r^-} + d_{r^-, v(r)}, & \text{if } ~~ r^- \in A^t  \\
		available(r^-) + d_{v^+(r^-), v(r)}, & \text{otherwise.}
	\end{cases}  \notag
\end{align}
If $r$ is the first request of its waiting vertex, the location $v(r)$ from which the vehicle travels towards request $r$ is necessarily the waiting vertex $\w(r)$. 
Otherwise, $v(r)$ depends on whether $r$ reveals by the time the vehicle finishes to satisfy the last accepted request: 
\begin{align}
v(r) 	&= \begin{cases}
		c_{r^-}, 		& \text{if } ~~ \Gamma_r \le  \max \big( available(r^-)+ d_{v^+(r^-), r^-}, e_{r^-} \big) + s_{r^-}  ~~ \wedge ~~ r^- \in A^t \\
		v(r^-), 	& \text{if } ~~ r^- \notin A^t   \\
		\w(r)  		& \text{otherwise.}
	\end{cases}  \notag
\end{align}

\paragraph{Request notifications.}

$A^t$ is the set of requests accepted up to time $t$. It is initialized with $A^{t-1}$ as all previously accepted requests must still be accepted at time $t$. Then incoming requests (\emph{i.e.} revealed at time $t$) are considered in increasing order with respect to $<_R$. $r$ is either accepted (added to $A^t$) or rejected (not added to $A^t$).
A request $r$ is accepted if (i) it is assigned to a waiting location, (ii) the vehicle is available, and (iii) its capacity is not exceeded. Formally, $r$ is added to $A^t$ if and only if:
\begin{equation}
\w(r)\neq\bot ~~ \wedge ~~ available(r) \le t^\text{max}_{r,v(r)} \wedge ~~ q_r + \sum_{r' \in \pi_k \cap A^t} q_{r'} \le Q .
\label{eq:conditions_R_plus}
\end{equation}

\paragraph{Vehicle operations.}

Once $A^t$ has been computed, vehicle operations for time unit $t$ must be decided. 
Each vehicle operates independently from all other vehicles.
If vehicle $k$ is traveling between a waiting vertex and a customer vertex, or if it is serving a request, then its operation remains unchanged. Otherwise, its operations are defined in Algorithm \ref{alg:veh}. 
\begin{algorithm}
\lIf{$t = \underline{on}(\s(w)) - d_{v, \s(w)}$}{travel from $v$ to $s(w)$}
\uElse{
	$P ~ \leftarrow$ set of requests of $\pi_w$ not yet satisfied, either accepted or not yet revealed\;
    \lIf{$P=\emptyset$}{travel to the next waiting vertex (or the depot)}
    \uElse{
    	$r^\text{next} ~ \leftarrow$ smallest element of $P$ according to the predefined order on $R$\;
        \uIf{$\w(r^\text{next}) = w$ \textbf{and} $r^\text{next}$ has already been revealed and accepted} 
        { wait until $t^\text{min}_{r^\text{next},v}$, travel to $r^\text{next}$ and satisfy the request; }
        \lIf{$\w(r^\text{next}) = w$ \textbf{but} $r^\text{next}$ is not known yet ($t < \Gamma_{r^\text{next}}$)}
        {travel back to waiting location $w$}
        \lIf{$\w(r^\text{next}) \ne w$}{wait until $\underline{on}\big(\s(w)\big) - d_{r, \s(w)}$ and travel to $\s(w)$}
    }
}
\protect\caption{Operations of vehicle $k$, at current time $t$.  
Vertex $v$ is the position of vehicle $k$, $w$ the waiting vertex it is currently assigned to, $\s(w)$ the waiting vertex (or the depot) that follows $w$ in $x$.
\label{alg:veh}}
\end{algorithm}


\subsubsection{Relation to strategy $\mathcal{R}^{\infty}$}

\begin{figure}
\centering
\includegraphics[width=1.0\textwidth]{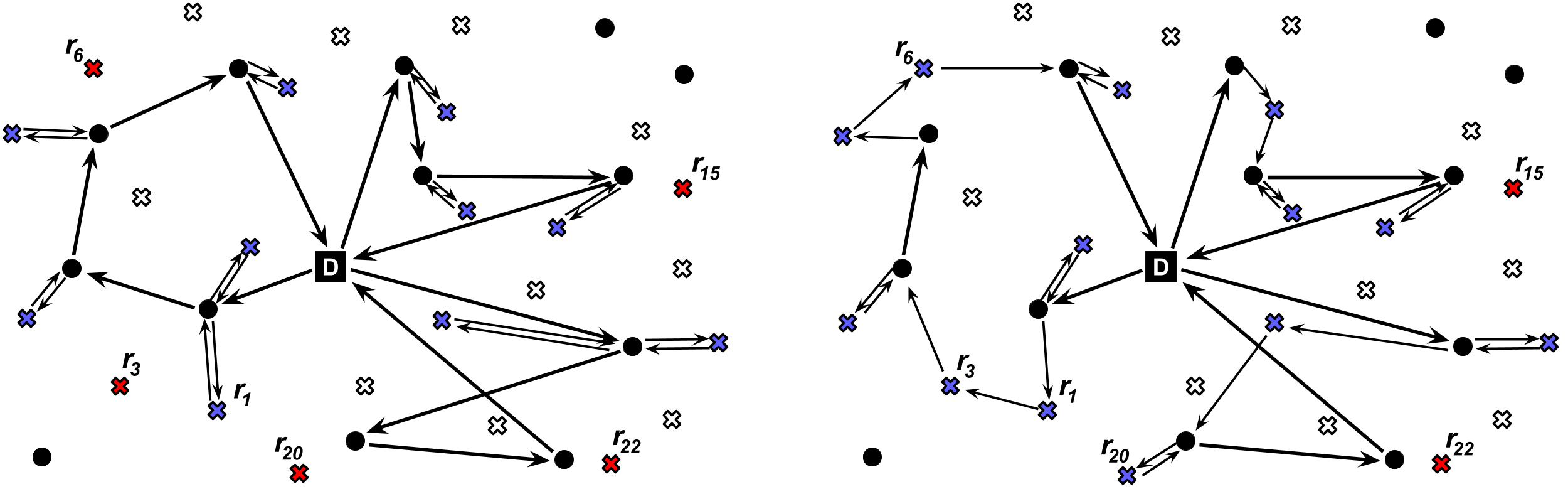}
\caption{
Comparative examples of strategies $\mathcal{R}^\infty$ (left) and $\mathcal{R}^{q+}$ (right). 
The depot, waiting vertices and customer vertices are represented by a square, circles and crosses, respectively. 
Arrows represent vehicle routes.
A filled cross represents a revealed request. 
Under $\mathcal{R}^\infty$, some requests ($r_3, r_6, r_{15}, \ldots$) can be missed. 
By avoiding pointless journeys when possible, $\mathcal{R}^{q+}$ is likely to end up with a lower number of missed requests. 
For example, if request $r_3$ is revealed by the time request $r_1$ is satisfied, then traveling directly to $r_3$ could help satisfy it. 
Similarly, on a different route, by traveling directly to the waiting vertex associated with request $r_{20}$, the vehicle could save enough time to satisfy $r_{20}$.
}
\label{fig:instance_strategies}
\end{figure}

In strategy $\mathcal{R}^{\infty}$ the vehicles handle requests by performing systematic round trips for their current waiting locations.
In $\mathcal{R}^{q+}$ , a vehicle travels directly towards a revealed request $r$ from a previously satisfied one $r$', provided that $r$ appears by the time the service of $r'$ gets completed. 
Furthermore, a vehicle is now allowed to travel directly from a customer vertex $c \in C$ to the next planned waiting vertex, without passing by the waiting vertex associated with $c$. 
Figure \ref{fig:instance_strategies} illustrates, informally, the differences between $\mathcal{R}^\infty$ and $\mathcal{R}^{q+}$.

\subsection{Expected cost of second-stage solutions under $\mathcal{R}^{q+}$ \label{sec:expected_cost_RQ_plus}}


Given a recourse strategy $\mathcal{R}$ and a first-stage solution $(x,\tau)$ to the SS-VRPTW-CR, 
a naive approach for computing $\mathcal{Q^R}(x,\tau)$ would be to follow the strategy described by $\mathcal{R}$ in order to confront $(x,\tau)$ with each and every possible scenario $\xi \subseteq R$.
Since there can be up to $2^{|R|}$ possible scenarios, this naive approach is not affordable in practice.
This section gives an overview of how we efficiently compute the expected number of rejected requests under the recourse strategy $\mathcal{R}^{q+}$. 
Further developments of the closed-form expressions are then provided in Appendix \ref{g1_Rplus}.

Recall that we assume that request probabilities are independent of each other; {\em i.e.}, for any couple of requests $r, r' \in R$, the probability $p_{r \wedge r'}$ that both requests will appear is given by $p_{r \wedge r'} = p_r \cdot p_{r'}$.
$\mathcal{Q}^{\mathcal{R}}(x,\tau)$ is equal to the expected number of rejected requests, which in turn is equal to the expected number of requests that are found to appear minus the expected number of accepted requests. Under the independence hypothesis, the expected number of revealed requests is given by the sum of all request probabilities, whereas the expected number of accepted requests is equal to the cumulative sum, for every request $r$, of the probability that it belongs to $A^h$, {\em i.e.}, 
\begin{equation}
\mathcal{Q}^{\mathcal{R}}(x,\tau) = \sum_{r \in R} p_r - \sum_{r \in R} \text{Pr}\{ r \in A^h \} 
= \sum_{r \in R} \big( p_r - \text{Pr}\{ r \in A^h \} \big) \label{eq:expectation}
\end{equation}

In the case of $\mathcal{R}^{q+}$, the satisfiability of a request $r$ depends on the current time and vehicle load, but also on the vertex from which the vehicle would leave to serve it.
The candidate vertices are necessarily either the current waiting location $w=\w(r)$ or any vertex hosting one of the previous requests associated with $w$.
Consequently, under $\mathcal{R}^{q+}$, the probability $\text{Pr}\{ r \in A^h \}$ is decomposed over all the possible time, load, and vertex configurations in which the vehicle can satisfy $r$:
\begin{align}
\text{Pr}\{ r \in A^h \} &= \sum_{t =t^\text{min+}_{r,w}}^{t^\text{max+}_{r,w}} \sum_{q=0}^{Q-q_r} g_1^{\w(r)}(r, t, q)
+ \sum_{\substack{r' \in \pi_w\\r' <_R r}} \sum_{t =t^\text{min+}_{r,r'}}^{t^\text{max+}_{r,r'}} \sum_{q=0}^{Q-q_r} g_1^{r'}(r, t, q) \label{eq:req-satisfiable-R_Q_plus} 
\end{align}
where: 
\begin{align}
g_1^{v}(r, t, q) \equiv \text{Pr}\{ & \text{request $r$ appeared} ~ \wedge ~ t = \max(t^\text{min}_{r, v}, available(r))  ~ \wedge \notag  \\
& v(r) = v ~ \wedge ~ \text{the vehicle carries a load of } q \} \notag 
\end{align}
Each tuple $(v,t, q)$ in the summation (\ref{eq:req-satisfiable-R_Q_plus}), where $v$ is either $\w(r)$ or a previously visited customer vertex $r'$, represents a possible configuration for accepting $r$. 
The probability to accept $r$ is then equivalent to the probability to fall into one of those states.
In particular, note that $\text{Pr}\{t = \max(t^\text{min}_{r, v}, available(r)) \}$ represents the probability that, if $r$ is accepted, the vehicle leaves its current position at time $t$ in order to satisfy it.
The calculus of $g_1^v$ is further developed in Appendix \ref{g1_Rplus}. 
Given $n$ customer vertices, a horizon of length $h$ and vehicle capacity of size $Q$, the computational complexity of computing the whole expected cost $\mathcal{Q}^{\mathcal{R}^{q+}}(x,\tau)$ is in $\mathcal{O}\big(n^2h^3Q \big)$, as detailed in the appendix.

\subsubsection*{Space complexity}
A naive implementation of equation (\ref{eq:req-satisfiable-R_Q_plus}) would basically fill up an $n^2 \times h^3 \times Q$ array. 
We draw attention to the fact that even a small instance with $n = Q = 10$ and $h=100$ would then lead to a memory consumption of $10^9$ floating point numbers. Using a common eight-byte representation requires more than seven gigabytes. 
Like strategy $\mathcal{R}^{\infty}$, important savings are obtained by noticing that the computation of $g_1$ functions for a given request $r$ under $\mathcal{R}^{q+}$ only relies on the previous potential request $r^-$. By computing $g_1$ while only keeping in memory the expectations of $r^-$ (instead of all $nh$ potential requests), the memory requirement is reduced by a factor $nh$. 
This however comes at the price of making any incremental computation, based on probabilities belonging to a similar first-stage solution, impossible.

\section{Progressive Focus Search for Static and Stochastic Optimization \label{sec:PFS}}

Solving a static stochastic optimization problem, such as the SS-VRPTW-CR, involves finding values for a set of first-stage decision variables that optimize an expected cost with respect to some recourse strategy:
$$ \min_{x} ~~ \mathcal{Q^R}(x), ~~~ x \in X $$
Solving this kind of problem is always challenging.
Besides the exponential size of the (first-stage) solution space $X$, the nature of the objective function $\mathcal{Q^R}$, an expectation, is usually computationally demanding. 
Because enumerating all possible scenarios is usually impossible in practice, some approaches tend to circumvent this bottleneck by restricting the set of considered scenarios, using for example the sample average approximation method (\citeauthor{ahmed2002sample}, \citeyear{ahmed2002sample}).
In some cases, expectations may be directly computed in (pseudo) polynomial time, by reasoning on the random variables themselves rather than on the scenarios. 
However, the required computational effort depends on the recourse strategy $\mathcal{R}$ and usually remains very demanding, as it is the case for the SS-VRPTW-CR. 

The Progressive Focus Search (PFS) metaheuristic aims at addressing these issues with two approximation factors, intended to reduce the size of the solution space and the complexity of the objective function. The initial problem $P_{\text{init}}$ is simplified into a problem $P_{\alpha,\beta}$ having simplified objective function and solution space. Parameters $\alpha$ and $\beta$ define the approximation factors of the objective function and of the solution space, respectively, and $P_{\alpha,\beta} = P_{\text{init}}$ when $\alpha = \beta = 1$. 
Whenever $\alpha > 1$ or $\beta > 1$, the optimal cost of $P_{\alpha,\beta}$ is an approximation of that of $P_{\text{init}}$. Starting from some initial positive values for $\alpha$ and $\beta$, the idea of PFS is to progressively decrease these values using an update policy. 
The simplified problem $P_{\alpha,\beta}$ is iteratively optimized for every valuation of $(\alpha,\beta)$, using the best solution found at the end of one iteration as starting point in the solution space for the next iteration.

The definition of the simplified problem $P_{\alpha,\beta}$ depends on the problem to be solved. In Sections \ref{sec:alpha} and \ref{sec:beta}, we give some general principles concerning $\alpha$ and $\beta$ and describe how to apply them to the case of the SS-VRPTW-CR. In Section \ref{sec:PFS_general}, we describe the generic PFS metaheuristic. 

\subsection{Reducing objective function computational complexity with $\alpha$ \label{sec:alpha}}

We assume the expected cost to be computed by filling matrices in several dimensions. 
In order to reduce the complexity, some of these dimensions must be scaled down.  
This is achieved by changing the scale of the input data and the decision variable domains related to the selected dimensions, dividing the values by the scale factor $\alpha$ and rounding to integer if necessary. 

For example, in the SS-VRPTW-CR the dimensions considered at computing the objective function are: the number of waiting vertices $n$, the vehicle capacity $Q$, and the time horizon $h$. Let $h=18000$ be the time horizon in the initial problem, corresponding to five hours in units of one second.
If we choose to reduce the time dimension with respect to a scale factor $\alpha=60$, then all durations in the input data (travel times, service times, time windows, etc.) are rounded to the nearest multiple of 60. Thus, the time horizon in the simplified problem $P_{\alpha,\beta} = P_{60,\beta}$ is of $h_{60}=300$, corresponding to a five-hour time horizon in units of one minute. The domains of waiting times decision variables are reduced accordingly, scaled from $[1,18000]$ in $P_{1,\beta}$ to $[1,300]$ in $P_{60,\beta}$.

Similarly, if we choose to reduce the vehicle capacity dimension with respect to a scale factor $\alpha=1000$, and if the vehicle capacity in the initial problem is $Q=500000$, \textit{e.g.} 500 kg in steps of 1 g, then all demands must be rounded to multiples of 1000. The capacity in $P_{1000,\beta}$ becomes $Q_{1000}=500$, thus 500 kg in units of 1 kg. 
When scaling dimensions of different nature, such as time and capacity, different scale factors should be considered, leading to a vector $\mathbf{\alpha}$.

Experiments have shown us that the closer $\alpha$ is to 1, the more accurate the approximation of the actual objective function is. 
Progressively reducing $\alpha$ during the search process allows us to quickly compute rough approximations at the beginning of the search process, when candidate solutions are usually far from being optimal, and spend more time computing more accurate approximations at the end of the search process, when candidate solutions get closer to optimality.

\subsection{Simplifying the solution space size with $\beta$ \label{sec:beta}}

When applying a scaling factor $\alpha$, for consistency reasons the nature of the scaled input data may impose to the domains of some decision variables to be reduced accordingly.
Yet the solution space can further simplified by reducing the domains of (part of) the remaining decision variables, or even by further reducing the same decision variables.
Let $Dom(v)$ be the initial set of values that may be assigned to $v$, that is, the domain of a decision variable $v$.
Domain reduction is not necessarily done for all decision variables, but only for a selected subset of them, denoted as $V_{\beta}$. The simplified problem is obtained by selecting $|Dom(v)|/\beta$ values and only considering these candidate values when searching for solutions, for each decision variable $v\in V_{\beta}$. 
Ideally, the selection of this subset of values should be done in such a way that the selected values are evenly distributed within the initial domain $Dom(v)$. 
We note $Dom_{\alpha,\beta}(v)$, the domain of a decision variable $v$ in the simplified problem $P_{\alpha,\beta}$.

For example, in the SS-VRPTW-CR a subset of decision variables defines the waiting times on the visited waiting vertices: $\tau_w$ defines the waiting time on $w$, with $Dom(\tau_w) = [1,h]$. 
If the temporal dimension is not scaled with respect to $\alpha$, or if $\alpha=1$, then $Dom_{\alpha,\beta}(\tau_w)$ is reduced to a subset of $[1,h]$ that contains $h/\beta$ values. To ensure that these values are evenly distributed in $[1,h]$, we may keep multiples of $\beta$. 
However, if the temporal dimension is scaled with respect to $\alpha$, the selected values must thereafter be scaled. 

Another subset of decision variables in the SS-VRPTW-CR defines the waiting vertices to be visited by the vehicles. The initial decision variable domains are then equivalent to $W$. Reducing the domains of these decision variables can be achieved by restricting to a subset of $W$ that contains $|W|/\beta$ waiting vertices. To ensure that these values are evenly distributed in the space, we may use geographical clustering techniques.

Progressively decreasing the value of $\beta$ allows us to progressively move from diversification to intensification: at the beginning of the search process, there are fewer candidate values for the decision variables of $V_{\beta}$. The solution method is therefore able to move quickly towards more fruitful regions of the search space. 
For minimization (resp. maximization) problems, we can easily show that the optimal solution of a simplified problem $P_{\alpha,\beta}$ is an upper (resp. lower) bound of the optimal solution of the problem $P_{\alpha,1}$; this is a direct consequence of the fact that every candidate solution of $P_{\alpha,\beta}$ is also a candidate solution of $P_{\alpha,1}$.

\subsection{PFS algorithm \label{sec:PFS_general}}

PFS requires the following input parameters: 
\begin{itemize}
\item An initial problem $P_{\text{init}}$;
\item Initial values $(\alpha_0,\beta_0)$ for $\alpha$ and $\beta$, as well as final values $(\alpha_\text{min},\beta_\text{min})$;
\item An update policy $\cal U$ that returns the new values $\alpha_{i+1}$ and $\beta_{i+1}$ given $\alpha_i$ and $\beta_i$; 
\item A computation time policy $\cal T$ such that ${\cal T}(\alpha,\beta)$ returns the  time allocated for optimizing $P_{\alpha,\beta}$;
\item A solution algorithm $\Theta$ such that, given a problem $P$, an initial solution $s$, and a time limit $\delta$, $\Theta(P,s,\delta)$ returns a possibly improved solution $s'$ for $P$.
\end{itemize}
\begin{algorithm}[b]
Initialize $i$ to 0 and construct an initial solution $s$ to problem $P_{init}$\;
\Repeat{$~\alpha_{i-1}=\alpha_\text{min} ~ \wedge ~ \beta_{i-1}=\beta_\text{min}$}{
	Build problem $P_{\alpha_i,\beta_i}$ and update the current solution $s$ to $P_{\alpha_i,\beta_i}$\;
	$s\leftarrow\Theta(P_{\alpha_i,\beta_i},s,{\cal T}(\alpha_i,\beta_i))$\;
	$(\alpha_{i+1},\beta_{i+1})\leftarrow {\cal U}(\alpha_{i},\beta_{i})$\;
	Increment $i$
}
\lIf{$\alpha_\text{min} > 1$}{Update the current solution $s$ to $P_{1,1}$}
\textbf{return} $s$\;
\protect\caption{Progressive Focus Search \label{alg:PFS-algo} (PFS)}
\end{algorithm}
PFS is described in Algorithm \ref{alg:PFS-algo}.
At each iteration $i$, the simplified problem $P_{\alpha_i,\beta_i}$ is built (line 3), and the current solution $s$ is updated accordingly (line 3): every value assigned to a decision variable which is concerned by the scale factor $\alpha$ is updated with respect to the new scale $\alpha_i$, and if a value assigned to a decision variable does not belong to the current domain associated with $\alpha_i$ and $\beta_i$, then it is replaced with the closest available value. 
Note that the updated solution may not be a feasible solution of $P_{\alpha_i,\beta_i}$ (because of value replacements and rounding operations on input data). Therefore the optimizer $\Theta$ must support starting with infeasible solutions.

Algorithm $\Theta$ is then used to improve $s$ with respect to the simplified problem $P_{\alpha_i,\beta_i}$ within a CPU time limit defined by the computation time policy ${\cal T}$ (line 4).
Finally, new values for $\alpha$ and $\beta$ are computed, according to the update policy ${\cal U}$ (line 5).
This iterative optimization process stops when $\alpha_{i-1}=\alpha_\text{min}$ and $\beta_{i-1}=\beta_\text{min}$, {\em i.e.}, when the last optimization of $s$ with $\Theta$ has been done with respect to the targeted level of accuracy defined by $(\alpha_\text{min},\beta_\text{min})$.
To ensure termination, we assume that the update policy $\cal U$ eventually returns $(\alpha_\text{min},\beta_\text{min})$ after a finite number of calls.
Finally, if the final value of $\alpha$ is larger than 1, so that $s$ is a scaled solution, then $s$ is scaled down to become a solution of the initial problem $P_{1,1}$ (line 8).

\section{Benchmark and Experimental Plan \label{sec:bench_experimental_plan}}

In this section, we introduce the new benchmark as well as the experimental concepts and tools used for experimentations reported in Sections \ref{sec:small_instances} and \ref{sec:big_instances}.

\subsection{A benchmark derived from real-world data \label{sec:instances}}

We derive our test instances from the benchmark described in \cite{Melgarejo2015} for the Time-Dependent TSP with Time Windows (TD-TSPTW). This benchmark has been created using real accurate delivery and travel time data obtained from the city of Lyon, France. 
It is available at
{\small \url{http://becool.info.ucl.ac.be/resources/ss-vrptw-cr-optimod-lyon}}, 
as well as the solution files and detailed result tables of the experiments conducted in the following sections.

The benchmark contains two different kinds of instances: instances with separated waiting locations and instances without separated waiting locations.
Each instance with separated waiting locations is denoted by $n$\textbf{c}-$m$\textbf{w}-$i$, where $n\in\{10,20,50\}$ is the number of customer vertices, $m\in \{5,10,30,50\}$ is the number of waiting vertices, and $x \in [1,15]$ is the random seed.
Each instance without separated waiting locations is denoted by $n$\textbf{c}+\textbf{w}-$i$. 
In these instances, every customer vertex is also a waiting vertex, $C = W$.
Instances sharing the same number of customers $n$ and the same random seed $x$ (\emph{e.g.} 50\textbf{c}-30\textbf{w}-1, 50\textbf{c}-50\textbf{w}-1 and 50\textbf{c+w}-1) always share exactly the same set of customer vertices $C$.
In all instances, the duration of an operational day is eight hours and the time horizon is $h=480$, which corresponds to one-minute time steps.

To each potential request $r=(c_r,\Gamma_r)$ is assigned a time window $[\Gamma_r, \Gamma_r+\Delta-1]$, where $\Delta$ is taken uniformly from $\{5,10,15,20\}$. Note that the time window always starts with the reveal time $\Gamma_r$.
This aims at simulating operational contexts similar to the practical application example described in introduction, the on-demand health care service at home, requiring immediate responses within small time windows. See the e-companion \ref{sec:instance_generation} for more details on the complete process used to generate instances.

\subsection{Compared approaches and experimental settings\label{sec:expe_plan}}

Experiments have been done on a cluster composed of 64-bit AMD Opteron 1.4-GHz cores.
The code is developed in \emph{C++11} with \emph{GCC4.9}, using \emph{-O3} optimization flag.
The current source code of our library for (SS-)VRPs is available from the online repository: {\small \url{bitbucket.org/mstguillain/vrplib}}.

We consider both recourse strategies $\mathcal{R}^{q+}$ and $\mathcal{R}^{q}$, a generalized version of $\mathcal{R}^{\infty}$ for capacitated vehicles. 
We compare their respective contribution and applicability, then we combine them to take the best of each, using several variations of PFS. 
An exact method allows us to measure optimality gaps, in order to assess the quality of the solutions found by PFS.
In order to evaluate the interest of exploiting stochastic knowledge, that is by modeling the problem as a SS-VRPTW-CR, the solutions are also compared with a wait-and-serve policy which does not anticipate, \textit{i.e.} in which vehicles are never relocated. 

\subsubsection{A capacitated version of $\mathcal{R}^{\infty}$ \label{sec:r_q}}

Recourse strategy $\mathcal{R}^{q+}$ is designed to be able to cope with vehicle maximal capacity constraints.
In order to compare both strategies $\mathcal{R}^{\infty}$ and $\mathcal{R}^{q+}$, and since part of our experiments involve limited vehicle capacities, an adapted version of $\mathcal{R}^{\infty}$ is required. 
We call this generalization $\mathcal{R}^{q}$. 
Vehicles behave under $\mathcal{R}^{q}$ exactly as under $\mathcal{R}^{\infty}$, but are limited by their capacity.
Its request acceptance rule follows the condition in  (\ref{eq:conditions_R_plus}), except that the definition of $available(r)$ and $t^\text{max}_{r,v(r)}$ are those stated in \cite{Saint-Guillain2017} for $\mathcal{R}^{\infty}$, and that $v(r) = \w(r)$.

In Appendix \ref{sec:R_Q_expectation}, we explain how to efficiently compute $\mathcal{Q}^{\mathcal{R}^{q}}(x,\tau)$. 
We also show how the resulting equations naturally reduce to the ones proposed in \cite{Bertsimas1992}, when particularized to the special case of the SS-VRP-C.
We found that, given $n$ customer vertices, a horizon of length $h$ and vehicle capacity of size $Q$, computing $\mathcal{Q}^{\mathcal{R}^{q}}(x,\tau)$ is of complexity $\mathcal{O}\big(nh^2Q \big)$. 
This is significantly lower than under $\mathcal{R}^{q+}$, which requires $\mathcal{O}\big(n^2h^3Q \big)$ operations in the worst case. 
However, such a lower complexity naturally comes at the price of a significantly higher expected cost in average, motivating the need for an adequate trade-off. 
We empirically address this question in Section \ref{sec:small_instances}.

\subsubsection{Progressive Focus Search.}

We have considered different update and computation time policies $\cal U$ and $\cal T$ in our experiments.
In this section, we only describe the optimizer $\Theta$ and the approximation factors $\alpha$ and $\beta$ used when conducting experiments with PFS.

\paragraph{Local Search Optimizer}
The optimizer $\Theta$ is the local search (LS) introduced in \cite{Saint-Guillain2017} to solve the SS-VRPTW-CR.
Starting from an initial randomly generated first-stage solution, LS iteratively modifies it by using a set of $9$ neighborhood operators: four classical ones for the VRP, {\em i.e.}, relocate, swap, inverted 2-opt, and cross-exchange (see \cite{kindervater1997vehicle,Taillard1997}), and five new operators dedicated to waiting vertices: insertion/deletion of a randomly chosen waiting vertex in/from $W^x$, increase/decrease of the waiting time $\tau_w$ of a randomly chosen vertex $\w \in W^x$, and transfer of a random waiting duration from one waiting vertex to another. 
After each modification of the first-stage solution, its expected cost is updated using the appropriate equations, depending on whether strategy $\mathcal{R}^{q}$ or $\mathcal{R}^{q+}$ is considered. 
The acceptance criterion follows the Simulated Annealing (SA) metaheuristic of \cite{Kirkpatrick1983}: improved solutions are always accepted, while degrading solutions are accepted with a probability which depends on the degradation and temperature. 
Temperature is initialized to $T_\text{init}$ and progressively decreased by a factor $f_T$ after each iteration of the LS. 
A restart strategy resets the temperature to its initial value each time it reaches a lower limit $T_\text{min}$. In all experiments, SA parameters were set to $T_\text{init}=2, T_\text{min} = 10^{-6}$, and $f_T = 0.95$.

\paragraph{Scale factor $\alpha$.}

In the initial problem $P_{1,1}$, temporal data is expressed with a resolution of one-minute time units.
The $\alpha$ factor is used to scale down this temporal dimension.
The time horizon is scaled down to $\mbox{round}(h/\alpha)$, so that each time step in $P_{\alpha,\beta}$ has a duration of $\alpha$ minutes. 
Every temporal input value (travel times $d_{i,j}$, reveal times $\Gamma_r$, service times $s_r$, and time windows $[e_r,l_r]$) is scaled from its initial value $t$ to $\mbox{round}(t/\alpha)$. 
Rounding operations are chosen in such a way that the desired quality of service is never underestimated by scaled data: $l_r$ is rounded down while all other values are rounded up. 
This ensures that a feasible first stage of a simplified problem $P_{\alpha, \beta}$ always remains feasible once adapted to $P_{1,1}$. 

\paragraph{Domain reduction factor $\beta$}

The decision variables concerned by domain reductions are waiting time variables: $V_\beta = \{\tau_w : w\in W\}$.
In $P_{1,1}$, we have $Dom(\tau_w) = [1,h]$.
Domains are reduced by selecting a subset of $|Dom(\tau_w)|/\beta$ values, evenly distributed in $[1,h]$. As the temporal dimension is also scaled with respect to $\alpha$, selected values are scaled down: $Dom_{\alpha,\beta}(\tau_w) = \{\mbox{round}(i/\alpha): i \in [1,h], i\mod \beta = 0\}$.

It is both meaningless (for vehicle drivers) and too expensive (for the optimization process) to design first-stage solutions with waiting times that are precise to the minute. Hence, in our experiments the domain of every waiting time decision variable is always reduced by a factor $\beta\geq 10$. When $\beta = 10$, waiting times are multiples of 10 minutes. When $\alpha=1$ and $\beta=10$, we have $Dom_{1,10}(\tau_w) = \{10, 20, 30, \ldots, 480\}$, but temporal data (travel and service times, time windows, \textit{etc.}) are precise to the minute.

\subsubsection{Enumerative exact method}

In order to assess the ability of our algorithms to find (\textit{near-}) optimal solutions, we devise a simple enumerative optimization method which is able to compute optimal solutions on small instances. 
To that end, the solution space is restricted to the solutions that \textit{(a)} use all available vehicles and \textit{(b)} use all the available waiting time.
Indeed, if $K \le |W|$, then on the basis of any optimal solution which uses only a subset of the available vehicles, a solution of the same cost can be obtained by assigning an idle vehicle to either a non-visited waiting vertex (if any) or the last visited vertex of any non-empty route (visiting at least two waiting locations), so that \textit{(a)} does not remove any optimal solution. 
Furthermore, if an optimal first-stage solution contains a route for which the vehicle returns to the depot before the end of the horizon, adding the remaining time to the last visited waiting vertex will never increase the expected cost of the solution, so that \textit{(b)} is also valid. 
The resulting solution space is then recursively enumerated in order to find the first-stage solution with the optimal expected cost.

\subsubsection{Wait-and-serve policy}

In order to assess the contribution of our recourse strategies, we compare them with a policy ignoring anticipative actions. 
This \emph{wait-and-serve} (w\&s) policy takes place as follows.
Vehicles begin the day at the depot. 
Whenever an online request $r$ appears, it is accepted if at least one of the vehicles is able to satisfy it, otherwise it is rejected.
If accepted, it is assigned to the \emph{closest such vehicle} which then visits it as soon as it becomes idle.
If there are several closest candidates, the least loaded vehicle is chosen.
After servicing $r$ (which lasts $s_r$ time units), the vehicle simply stays idle at $r$'s location until it is assigned another request or until it must return to the depot. 
Note that a request cannot be assigned to a vehicle if satisfying it prevents the vehicle from returning to the depot before the end of the horizon.

Note that, whereas our recourse strategies for the SS-VRPTW-CR generalize to requests such that the time window starts later than the reveal time, in our instances we consider only requests where $e_r = \Gamma_r$.
Doing it the other way would in fact require a more complex wait-and-serve policy, since the current version would be far less efficient and unrealistic in the case of requests with $e_r$ significantly greater than $\Gamma_r$.

In what follows, average results are always reported for the w\&s policy. We randomly generate $10^6$ scenarios according to the $p_r$ probabilities. For each scenario, we apply the w\&s approach to compute a number of rejected requests; finally, the average number of rejected requests is reported. 
The results of PFS and the exact method are always reported by means of average relative gains, in percentages, with respect to the w\&s policy: the gain of a first-stage solution $s$ computed with PFS or the exact method is $\frac{\text{avg}-E}{\text{avg}}$, where $E$ is the expected cost of $s$ and $avg$ is the average cost under the w\&s policy.

\section{Experiments on small instances \label{sec:small_instances}}

We consider small test instances, having $n \in \{10,20\}$ customer vertices. 
Furthermore, PFS is here instantiated such that we perform only a single optimization step (lines 2-7 of Algorithm \ref{alg:PFS-algo}): $\alpha_0 = \alpha_\text{min}$ and $\beta_0 = \beta_\text{min}$. 
The simplified problem $P_{\alpha,\beta}$ is therefore first optimized for a duration of $T$ seconds, and the returned solution is adapted with respect to the initial problem $P_{1,1}$, ensuring that all results are expressed according to the original input data. 
This limited experimental setting, while ignoring the impact of performing several optimization steps in PFS, aims at determining:
\begin{enumerate}

\item Whether the loss of precision, introduced by $\alpha$ and $\beta$, is counterbalanced by the fact that the approximation $P_{\alpha,\beta}$ is easier to solve than the initial problem.

\item The impact of avoiding pointless trips in recourse strategy $\mathcal{R}^{q+}$, compared with simpler (but computationally less demanding) strategy $\mathcal{R}^{q}$. 

\item The interest of exploiting stochastic knowledge, by comparing the expected costs of SS-VRPTW-CR solutions with their average costs under the w\&s policy.

\item The quality of the solutions computed by the LS algorithm under different scale factors. 
These are compared with optimal solutions obtained with the exact method. When $\alpha>1$ or $\beta>1$, the exact method solves $P_{\alpha,\beta}$, and the results are reported according to the final solution, scaled back to $P_{1,1}$. 
\end{enumerate}

\subsection{Impact of the scale factor $\alpha$}

\begin{table}
\footnotesize
\centering
\begin{tabular}{l  r r    r       c   r         r      c   r         r       c   r               r       c  r               r     c  r                r          }

          &       & \multicolumn{8}{c}{Exact (\% gain after 30 minutes)}                  &   & \multicolumn{8}{c}{PFS (\% gain after 5 minutes)}    \\
                           \cmidrule{3-10}                                                      \cmidrule{12-19}                              
          &       & \multicolumn{2}{c}{$\alpha = 1$} && \multicolumn{2}{c}{$\alpha = 2$} && \multicolumn{2}{c}{$\alpha = 5$} && \multicolumn{2}{c}{$\alpha = 1$} && \multicolumn{2}{c}{$\alpha = 2$} && \multicolumn{2}{c}{$\alpha = 5$} \\
 \cmidrule{3-4} \cmidrule{6-7} \cmidrule{9-10}             \cmidrule{12-13}     \cmidrule{15-16} \cmidrule{18-19}                          
 			& w\&s  & $\mathcal{R}^{q}$     & $\mathcal{R}^{q+}$    && $\mathcal{R}^{q}$       & $\mathcal{R}^{q+}$     && $\mathcal{R}^{q}$       & $\mathcal{R}^{q+}$      && $\mathcal{R}^{q}$      & $\mathcal{R}^{q+}$     && $\mathcal{R}^{q}$     & $\mathcal{R}^{q+}$     && $\mathcal{R}^{q}$      & $\mathcal{R}^{q+}$       \\ \hline
10c-5w-1  & 12.8 & 8.9*   &\bf 15.4  && 7.3*    & 12.8*  &~& 4.4*  & 9.5*   		&& 8.9     & 14.1   && 7.3     & 13.1   && 4.4       & 9.2   \\
10c-5w-2  & 10.8 & -4.8*  &\bf 7.4   && -4.8*   &\bf 7.4*   && -8.8*  & 0.5*   		&& -4.8    &  0.2   &&  -4.8   &  4.9   && -8.8      & 0.5   \\
10c-5w-3  & 8.0  & -46.9* &\bf -26.5 &~& -46.9* &\bf -26.5* && -55.9* & -43.2* 		&& -46.9   & -29.9  && -43.1   & -30.6  && -43.1     & -32.9 \\
10c-5w-4  & 10.5 & -10.9* &\bf 0.9   && -10.9*  &\bf 0.9*   && -10.9* &\bf 0.9*   		&& -10.9   & -8.7   && -10.9   & -4.9   && -10.9     & -2.2  \\
10c-5w-5  & 8.4  & -17.9* &\bf 2.5   &~& -17.9* &\bf 2.5    && -20.5* & 0.5*   		&& -17.9   & -6.1   && -18.9   & -2.8   && -19.5     & 1.1   \\ \hline
\#eval 	  &      &  $10^4$& &&$10^4$ &  && $10^4$ & $10^4$		&&$3^*10^4$&$3^*10^3$&&$7^*10^4$&$5^*10^3$&&$2^*10^5$&$2^*10^4$  \\ \hline

10c+w-1  & 12.8  & 35.3 	& 34.4  && 35.3    & 34.4   && 32.7   & 26.5      	&&\bf 39.1    & 30.3   && 38.3    & 36.7   && 34.9     & 34.1  \\
10c+w-2  & 10.8  & 28.1 	& 19.1  && 30.1    & 21.5   && 30.1   & 29.7      	&&\bf 32.3    & 18.8   && 32.1    & 25.2   &&\bf 32.3     & 25.8  \\
10c+w-3  & 8.0   & 14.4 	& 17.1  && 18.8    & 17.1   && 13.3   & 13.7     	&& 26.1    & 18.8   &&\bf 27.6    & 20.3   && 23.1     & 23.9  \\
10c+w-4  & 10.5  & 7.8  	& 11.0  && 12.4    & 11.6   && 7.8    & 11.4      	&& 22.6    & 12.3   &&\bf 23.3    & 16.4   && 18.8     & 16.5  \\
10c+w-5  & 8.4   & 3.5  	& 8.9   && 8.4     & 6.6    && 23.7   & 1.7       	&& 31.6    & 15.8   &&\bf 32.7    & 21.1   && 29.3     & 28.5  \\ \hline
\#eval   &       & &&& &&&&  &&$3^*10^4$&$3^*10^3$&& $7^*10^4$&$6^*10^3$&&$2^*10^5$&$2^*10^4$  \\ \hline
\end{tabular}
\caption{Results on small instances ($n=10$, $K=2$, $Q = \infty$) when $\alpha\in\{1,2,5\}$ and $\beta=60$.
For each instance, we give the average cost over $10^6$ sampled scenarios using the \emph{wait-and-serve} policy (\textit{w\&s}) and the gain of the best solution found by the exact approach within a time limit of 30 minutes and PFS within a time limit of 5 minutes (average on 10 runs). 
Results marked with a star ($^*$) have been proved optimal.
\#eval gives the average number of expectation computations for each run: solutions enumerated (Exact) or LS iterations (PFS).
} 
\label{table:results_2veh_infty}
\end{table}

Table \ref{table:results_2veh_infty} shows the average gains, in percentages, of using an SS-VRPTW-CR solution instead of the w\&s policy, for small instances composed of $n=10$ customer vertices with $K=2$ uncapacitated vehicles. 
We consider three different values for $\alpha$. When $\alpha=1$ (resp. $\alpha=2$, $\alpha=5$), the time horizon is $h=480$ (resp. $h_\alpha=240$, $h_\alpha=96$) and each time unit corresponds to one minute (resp. two and five minutes).
In all cases, the domain reduction factor $\beta$ is set to $60$: waiting times are restricted to multiples of 60 minutes.

Unlike the recourse strategies, which must to deal with a limited set of predefined waiting locations, the w\&s policy makes direct use of the customer vertices.
Therefore, the relative gain of using an optimized SS-VRPTW-CR first-stage solution is highly dependent on the locations of the waiting vertices. 
Gains are always greater for \emph{10c+w-$i$} instances, where any customer vertex can be used as a waiting vertex: for these instances, gains with the best-performing strategy are always greater than $23\%$, whereas for {\em 10c-5w-i} instances, the largest gain is $16\%$, and is negative in some cases.

The results obtained on instance \emph{10c-5w-3} are quite interesting: gains are always negative; {\em i.e.}, waiting strategies always lead to higher expected numbers of rejected requests than the w\&s policy. 
By looking further into the average travel times in each instance, in Table \ref{table:avg_travel_times_10c}, we find that the average travel time between customer vertices in instance \emph{10c-5w-3} is rather small (12.5), and very close to the average duration of time windows (12.3). In this case, anticipation is of less importance and the w\&s policy appears to perform better. Furthermore, average travel time between waiting and customer vertices (19.5) is much larger than the average travel time between customer vertices. 

\begin{table}
\small
\centering
\begin{tabular}{@{}lrrrrr@{}}
\textit{Instance:}    & 10c-5w-1 & 10c-5w-2 & 10c-5w-3 & 10c-5w-4 & 10c-5w-5 \\ \midrule
\textit{Travel time within $C$:}          & 19.6' & 16.8'  & 12.5' & 18.0' & 13.0'           \\
\textit{Travel time between $C$ and $W$:} & 23.7'   & 19.9'  & 19.5' & 20.9' & 18.2'           \\
\textit{Time window duration:}               & 11.6' & 12.7'  & 12.3' & 13.2' & 12.6'         \\\hline
\end{tabular}
\caption{Statistics on instances 10c-5w-$i$: the first (resp. second) line gives the average travel time between customer vertices (resp. between a customer and waiting vertices); the last line gives the average duration of a time window.
\label{table:avg_travel_times_10c}}
\end{table}

We note that the exact enumerative method runs out of time under $\mathcal{R}^{q+}$ for all instances, when $\alpha = 1$. 
Increasing $\alpha$ to 2 speeds up the solution process and makes it possible to prove optimality on all \emph{10c-5w-$i$} instances except instance 5. Setting $\alpha=5$ allows to find all optimal solutions.
However, optimizing with coarser scales may degrade the solution quality.
This is particularly true for \emph{10c-5w-$i$} instances which are easier, in terms of solution space, than {\em 10c+w-x} instances as they have half the number of waiting locations: for \emph{10c-5w-$i$} instances, gains are often decreased when $\alpha$ is increased because, whatever the scale is, the search finds optimal or near-optimal solutions. 

For PFS, gains with recourse strategy $\mathcal{R}^{q+}$ are always greater than gains with recourse strategy $\mathcal{R}^{q}$ on \emph{10c-5w-$i$} instances. 
However, we observe the opposite on {\em 10c+w-i} instances.
This comes from the fact that expected costs are much more expensive to compute under $\mathcal{R}^{q+}$ than under $\mathcal{R}^{q}$. 
Table \ref{table:results_2veh_infty} displays the average number of times the objective function $\mathcal{Q}^{\mathcal{R}}(x,\tau)$ is evaluated (\#eval), that is the number of solutions considered by either the local search or the exact method, in which case it corresponds to the size of the solution space (when enumeration is complete and under assumptions \textit{(a)} and \textit{(b)} discussed in Section \ref{sec:expe_plan}). 
We note that the number of LS iterations is $\pm$10 times smaller when using $\mathcal{R}^{q+}$ compared to $\mathcal{R}^{q}$. 
As \emph{10c-5w-$i$} instances are easier than {\em 10c+w-i} instances, around $10^4$ iterations is enough to allow the LS optimizer of PFS to find near-optimal solutions. In this case, gains obtained with $\mathcal{R}^{q+}$ are much larger than those obtained with $\mathcal{R}^{q}$. However, on {\em 10c+w-i} instances, $10^4$ iterations are not enough to find near-optimal solutions. For these instances, better results are obtained with $\mathcal{R}^{q}$.

When optimality has been proven by Exact, we note that PFS often finds solutions with the same gain.
With $\alpha\in\{2,5\}$, PFS may even find better solutions: this is due to the fact that optimality is only proven for the simplified problem $P_{\alpha,\beta}$, whereas the final gain is computed after scaling back to the original horizon at scale 1. 
When optimality has not been proven, PFS often finds better solutions (with larger gains).

\subsection{Combining recourse strategies: $\mathcal{R}^{q/q+}$}

Results obtained from Table \ref{table:results_2veh_infty} show that although it leads to larger gains, the computation of expected costs is much more expensive under recourse strategy $\mathcal{R}^{q+}$ than under $\mathcal{R}^{q}$, which eventually penalizes the optimization process as it performs fewer iterations within the same time limit (for both Exact and PFS). 

We now introduce a pseudo-strategy that we call $\mathcal{R}^{q/q+}$, which combines $\mathcal{R}^{q}$ and $\mathcal{R}^{q+}$. 
For both Enum and PFS, strategy $\mathcal{R}^{q/q+}$ refers to the process that uses $\mathcal{R}^{q}$ as the evaluation function during all the optimization process. When stopping at a final solution, we reevaluate it using $\mathcal{R}^{q+}$.
Table \ref{table:results_2veh_inftybis} reports the gains obtained by applying $\mathcal{R}^{q/q+}$ on instances \emph{10c-5w-$i$} and \emph{10c+w-$i$}.
By using $\mathcal{R}^{q/q+}$, we actually use $\mathcal{R}^{q}$ to guide the LS optimization, which permits the algorithm to consider a significantly bigger part of the solution space.
For both Enum and PFS, $\mathcal{R}^{q/q+}$ always leads to better results than $\mathcal{R}^{q}$. 
From now on, we will only consider strategies $\mathcal{R}^{q/q+}$ and $\mathcal{R}^{q+}$ in the next experiments.

\begin{table}
\footnotesize
\centering
\begin{tabular}{l c r c r    r       c   r         r      c   r         r       c   r               r       c  r               r     c  r                r          }

         & &       & & \multicolumn{8}{c}{Exact (\% gain after 30 minutes)}                  &   & \multicolumn{8}{c}{PFS (\% gain after 5 minutes)}    \\
                           \cmidrule{5-12}                                                      \cmidrule{14-21}                              
         & &       & & \multicolumn{2}{c}{$\alpha = 1$} && \multicolumn{2}{c}{$\alpha = 2$} && \multicolumn{2}{c}{$\alpha = 5$} && \multicolumn{2}{c}{$\alpha = 1$} && \multicolumn{2}{c}{$\alpha = 2$} && \multicolumn{2}{c}{$\alpha = 5$} \\
 \cmidrule{5-6} \cmidrule{8-9} \cmidrule{11-12}             \cmidrule{14-15}     \cmidrule{17-18} \cmidrule{19-21}                          
 && w\&s  && $\mathcal{R}^{q}$     & $\mathcal{R}^{q/q+}$    && $\mathcal{R}^{q}$       & $\mathcal{R}^{q/q+}$     && $\mathcal{R}^{q}$       & $\mathcal{R}^{q/q+}$      && $\mathcal{R}^{q}$      & $\mathcal{R}^{q/q+}$     && $\mathcal{R}^{q}$     & $\mathcal{R}^{q/q+}$     && $\mathcal{R}^{q}$      & $\mathcal{R}^{q/q+}$       \\ \hline
10c-5w-1  && 12.8 && 8.9    &\bf 14.0  && 7.3     & 12.8   &~& 4.4   & 9.5    		&& 8.9     &\bf 14.0   && 7.3     & 12.8   && 4.4       & 9.5   \\
10c-5w-2  && 10.8 && -4.8   &\bf 7.4   && -4.8    &\bf 7.4    && -8.8   & 0.5    		&& -4.8    &  7.1   &&  -4.8   &\bf  7.4   && -8.8      & 0.5   \\
10c-5w-3  && 8.0  && -46.9  & -26.5 &~& -46.9  & -26.5  && -55.9  & -37.3  		&& -46.9   & -26.5  && -43.1   &\bf -22.5  && -43.1     &\bf -22.5 \\
10c-5w-4  && 10.5 && -10.9  &\bf 0.9   && -10.9   &\bf 0.9    && -10.9  &\bf 0.9    		&& -10.9   &\bf 0.9    && -10.9   &\bf  0.9   && -10.9     &\bf 0.9  \\
10c-5w-5  && 8.4  && -17.9  &\bf 2.5   &~& -17.9  &\bf 2.5    && -20.5  & 0.5    		&& -17.9   &\bf 2.5    && -18.9   &  1.8   && -19.5     & 1.1   \\ \hline
10c+w-1   && 12.8 && 35.3 	& 37.7  && 35.3    & 37.7   && 32.7   & 34.0      	&& 39.1    &\bf 40.9   && 38.3    & 39.5   && 34.9     & 36.2  \\
10c+w-2   && 10.8 && 28.1 	& 33.2  && 30.1    & 34.6   && 30.1   & 34.6      	&& 32.3    &\bf 35.6   && 32.1    & 35.5   && 32.3     & 34.4  \\
10c+w-3   && 8.0  && 14.4 	& 24.4  && 18.8    & 29.9   && 13.3   & 21.9     	&& 26.1    & 35.0   && 27.6    &\bf 35.6   && 23.1     & 30.3  \\
10c+w-4   && 10.5 && 7.8  	& 13.5  && 12.4    & 20.0   && 7.8    & 13.5      	&& 22.6    & 29.2   && 23.3    &\bf 29.3   && 18.8     & 24.1  \\
10c+w-5   && 8.4  && 3.5  	& 10.6  && 8.4     & 13.3   && 23.7   & 31.0       	&& 31.6    & 39.8   && 32.7    &\bf 40.7   && 29.3     & 35.0  \\ \hline
\end{tabular}
\caption{Comparison of $\mathcal{R}^{q}$ with the hybrid strategy $\mathcal{R}^{q/q+}$ (that uses strategy $\mathcal{R}^{q}$ as evaluation function during the optimization process, and evaluates the final solution with strategy $\mathcal{R}^{q+}$) on the small instances used in Table \ref{table:results_2veh_infty}, with $\beta=60$.} 
\label{table:results_2veh_inftybis}
\end{table}

\subsection{Impact of the domain reduction factor $\beta$}

Table \ref{table:results_10min_LS} considers instances involving 20 customer vertices and either 10 separated waiting locations (\emph{20c-10w-$i$}) or one waiting location at each customer vertex (\emph{20c+w-$i$}).
It compares results obtained by PFS for two different computation time limits, with $\beta\in\{10,30,60\}$. When $\beta=10$ (resp. $\beta=30$ and $\beta=60$), domains of waiting time variables contain 48 (resp. 16 and 8) values, corresponding to multiples of 10 (resp. 30 and 60) minutes. In all cases, the scale factor $\alpha$ is set to 2.

When considering the recourse strategy ${\cal R}^{q+}$ with a five-minute computation time limit, we observe that better results are obtained with $\beta=60$, as domains are much smaller. When the computation time is increased to 30 minutes, or when considering strategy ${\cal R}^{q/q+}$, which is cheaper to compute, then better results are obtained with $\beta=10$, as domains contain finer-grained values.

We observe that $\mathcal{R}^{q/q+}$ always provides better results than pure $\mathcal{R}^{q+}$, whatever the waiting time multiple $\beta$ used. Except when switching to significantly greater computational times, $\mathcal{R}^{q/q+}$ seems more adequate as it combines the limited computational cost incurred by $\mathcal{R}^{q}$ with the nicer expected performances of the cleverer strategy $\mathcal{R}^{q+}$.

\begin{table}
\footnotesize
\centering
\begin{tabular}{l c r c r    r       c   r         r      c   r         r       c   r               r       c  r               r     c  r                r          }

         & &       & & \multicolumn{8}{c}{Exact (\% gain after 30 minutes)}                  &   & \multicolumn{8}{c}{PFS (\% gain after 5 minutes)}    \\
                           \cmidrule{5-12}                                                      \cmidrule{14-21}                              
         & &       & & \multicolumn{2}{c}{$\beta = 60$} && \multicolumn{2}{c}{$\beta = 30$} && \multicolumn{2}{c}{$\beta = 10$} && \multicolumn{2}{c}{$\beta = 60$} && \multicolumn{2}{c}{$\beta = 30$} && \multicolumn{2}{c}{$\beta = 10$} \\
 \cmidrule{5-6} \cmidrule{8-9} \cmidrule{11-12}             \cmidrule{14-15}     \cmidrule{17-18} \cmidrule{19-21}                          
 && w\&s  && $\mathcal{R}^{q/q+}$     & $\mathcal{R}^{q+}$    && $\mathcal{R}^{q/q+}$       & $\mathcal{R}^{q+}$     && $\mathcal{R}^{q/q+}$       & $\mathcal{R}^{q+}$      && $\mathcal{R}^{q/q+}$      & $\mathcal{R}^{q+}$     && $\mathcal{R}^{q/q+}$     & $\mathcal{R}^{q+}$     && $\mathcal{R}^{q/q+}$      & $\mathcal{R}^{q+}$       \\ \hline
20-c10w-1 && 22.6 && 9.3  & -12.1&& 9.7   &-11.4 &&\bf 12.6  & -16.0 		 && 10.8  &  5.5  && 12.0  &  5.6 &&\bf 15.2 & 6.4  \\
20-c10w-2 && 19.8 && -11.8& -27.9&& -5.0  &-29.1 &&\bf -3.7  & -31.4 		 && -5.7  & -12.2 && -3.8  &-10.2 &&\bf -1.9 & -8.7  \\
20-c10w-3 && 21.1 && -0.5 & -16.7&&  5.6  &-15.9 &&\bf  6.4  & -21.3 		 &&  1.1  & -2.8  &&  7.7  & -0.9 &&\bf 8.1  & 0.2   \\
20-c10w-4 && 25.3 && 4.6  & -4.3 &&  5.2  & -6.9 &&\bf  5.4  &  -9.2  		 &&\bf  5.7  &  4.3  &&  5.5  &  3.8 && 5.1  & 4.7   \\
20-c10w-5 && 20.9 && -10.7& -25.9&&  -1.0 &-24.6 &&\bf  0.2  & -23.8  		 && -9.1  &-14.0  && -0.0  & -7.0 &&\bf 0.8  & -6.4  \\
\hline
20-c+w-1 && 22.6 && 15.4 & 2.8  && 17.2  &  2.2 &&\bf 17.4  &  0.2   		&& 17.9  & 13.5  && 19.4  & 11.4 &&\bf 20.2 & 13.0  \\
20-c+w-2 && 19.8 &&\bf 7.6  & -10.0&&  5.6  &-16.8 && 6.9   & -14.3  		&& 12.2  &  2.7  && 10.7  &  2.1 &&\bf 12.3 & 3.5  \\
20-c+w-3 && 21.1 && 2.8  & -11.1&&\bf  3.9  &-14.1 &&  3.1  & -13.7  		&&  4.8  &  1.2  &&  6.4  &  0.4 &&\bf 7.8  & 0.3  \\
20-c+w-4 && 25.3 && 14.3 & 5.2  &&\bf 15.3  &  2.6 && 14.0  &  3.0   		&& 15.9  & 12.4  && 18.6  & 13.5 &&\bf 19.6 & 14.2  \\
20-c+w-5 && 20.9 && 13.6 & -2.6 && 14.0  &-11.2 &&\bf 16.7  & -7.7   		&& 15.7  &  9.8  &&\bf 19.0  & 11.0 && 18.5 & 12.0  \\ 
\hline
\end{tabular}
\caption{Relative gains 5 and 30 minutes, using three domain reduction factors ($\beta\in\{10,30,60\}$), with $K=2$ uncapacitated vehicles and a scale factor $\alpha=2$. Instances involve $n=20$ customer locations and either 10 or 20 available waiting locations.} 
\label{table:results_10min_LS}
\end{table}

\section{Experiments on large instances \label{sec:big_instances}}

We now consider instances with $n=50$ customer vertices. Instances \emph{50c-30w-$i$} and \emph{50c-50w-$i$} have $m=30$ and $m=50$ separated waiting locations, respectively. Instances \emph{50c+w-$i$} have $m=50$ waiting vertices which correspond to the customer vertices. Each class is composed of 15 instances such that, for each seed $i\in [1,15]$, the three instances classes \emph{50c-30w-i}, \emph{50c-50w-i}, and \emph{50c+w-i} contain the same set of 50 customer vertices and thus only differ in terms of the number and/or positions of waiting vertices.
For each instance, the vehicle's capacity is set to $Q=20$, and we consider three different numbers of vehicles $K\in\{5, 10, 20\}$.
In total, we thus have 45 $\times$ 3 = 135 different configurations. 

We first compare and discuss the behaviors of different instantiations of PFS. 
Then, based on the PFS variant that appears to perform best, further experiments (Section \ref{sec:big_instances_results}) measure the contribution of a two-stage stochastic model, through the use of a SS-VRPTW-CR formulation and our recourse strategies. 

\subsection{Instantiations of PFS}

All runs of PFS are limited to $T=10800$ seconds (three hours).
We compare seven instantiations of PFS, which have different update and computation time policies $\cal U$ and $\cal T$, while all other parameters are set as described in Section 6.2.1. 
Strategy $\mathcal{R}^{q/q+}$ is used for all experiments. 
The different instantiations are:
\begin{itemize}

\item \emph{PFS-$\alpha$*$\beta$10}: the scale factor $\alpha$ is progressively decreased from 5 to 2 and 1 while the domain reduction factor $\beta$ remains fixed to $10$. 
More precisely, $\alpha_0=5$, $\alpha_\text{min}=1$, and $\beta_0=\beta_\text{min}=10$. The update policy $\cal U$ successively returns $\alpha_1=2$ and $\alpha_2=1$, while $\beta_1=\beta_2=10$. 
The computation time policy $\cal T$ always returns 3600 seconds, so that the three LS optimizations have the same CPU time limit of one hour.

\item \emph{PFS-$\alpha$1$\beta$*}: $\alpha$ remains fixed to 1 while $\beta$ is progressively decreased from 60 to 30 and 10. 
More precisely, $\alpha_0=\alpha_\text{min}=1$, $\beta_0=60$, and $\beta_\text{min}=10$. 
The update policy $\cal U$ successively returns $\beta_1=30$ and $\beta_2=10$, while $\alpha_1 = \alpha_2 = 1$. 
The computation time policy $\cal T$ always returns 3600 seconds.

\item \emph{PFS-$\alpha$*$\beta$*}: both $\alpha$ and $\beta$ are progressively decreased. 
We set $\alpha_0=5$, $\alpha_\text{min}=1$, $\beta_0=60$, and $\beta_\text{min}=10$. 
The update policy $\cal U$ returns the following couples of values for $(\alpha_i,\beta_i)$: (2, 60), (1, 60), (5, 30), (2, 30), (1, 30), (5, 10), (2, 10), (1, 10). 
The computation time policy $\cal T$ always returns 1200 seconds. 
The PFS optimization process is hence composed of nine LS optimizations of 20 minutes each.

\item \emph{PFS-$\alpha$a$\beta$b} which performs only a single LS optimization step with $T=10800$ and $\alpha_0 = \alpha_\text{min} = a$ and $\beta_0 = \beta_\text{min} = b$, as experimented in Section 7. We consider two different values for $\alpha$, {\em i.e.}, $a \in \{1,2\}$, and two different values for $\beta$, {\em i.e.}, $b \in \{10,60\}$, thus obtaining four different instantiations.

\end{itemize}

\subsection{Comparison of the different PFS instantiations}

\begin{figure}
\centering
\begin{minipage}[c]{.47\linewidth}
      \includegraphics[width=1.0\textwidth]{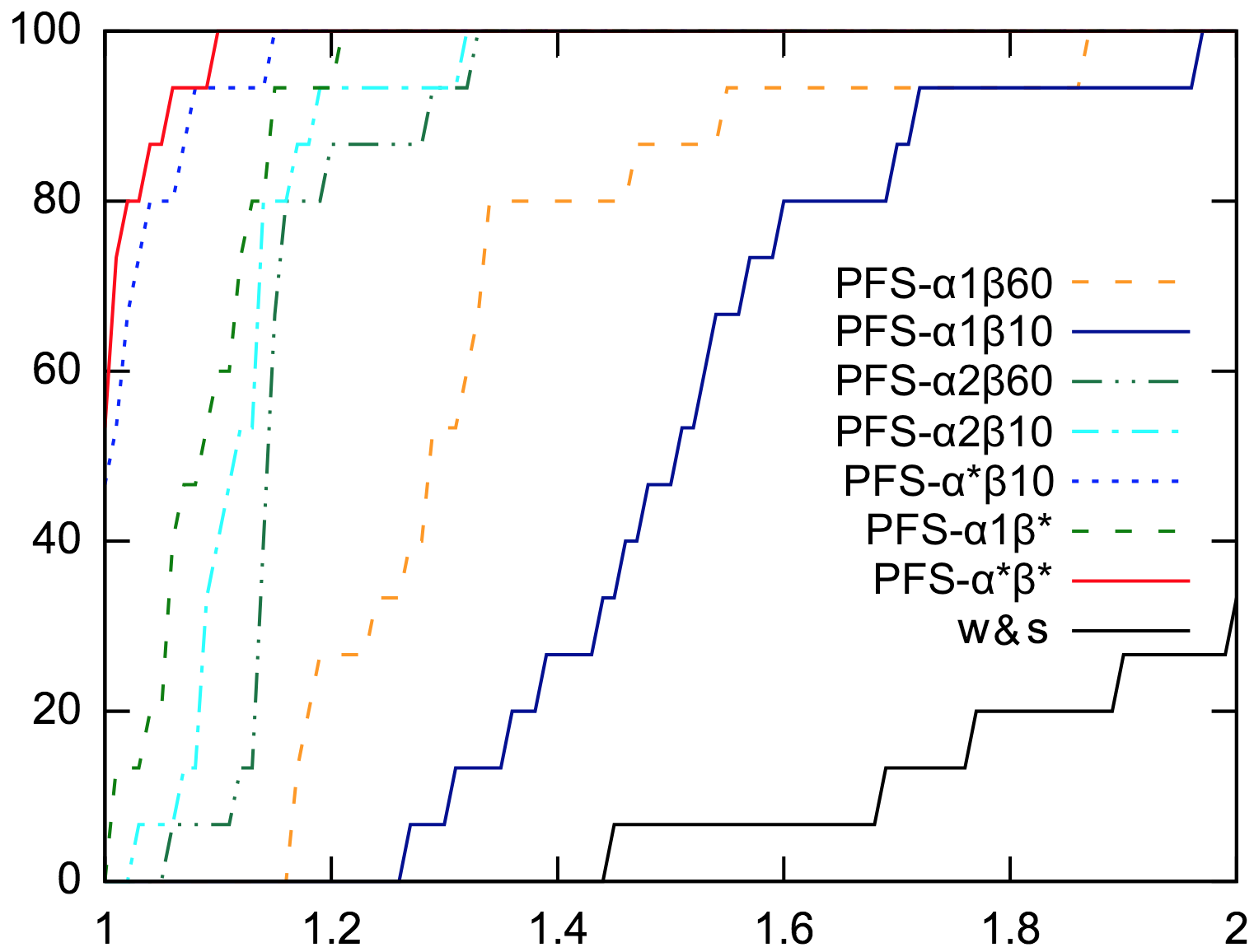}
   \end{minipage} \hfill
   \begin{minipage}[c]{.47\linewidth}
      \includegraphics[width=1.0\textwidth]{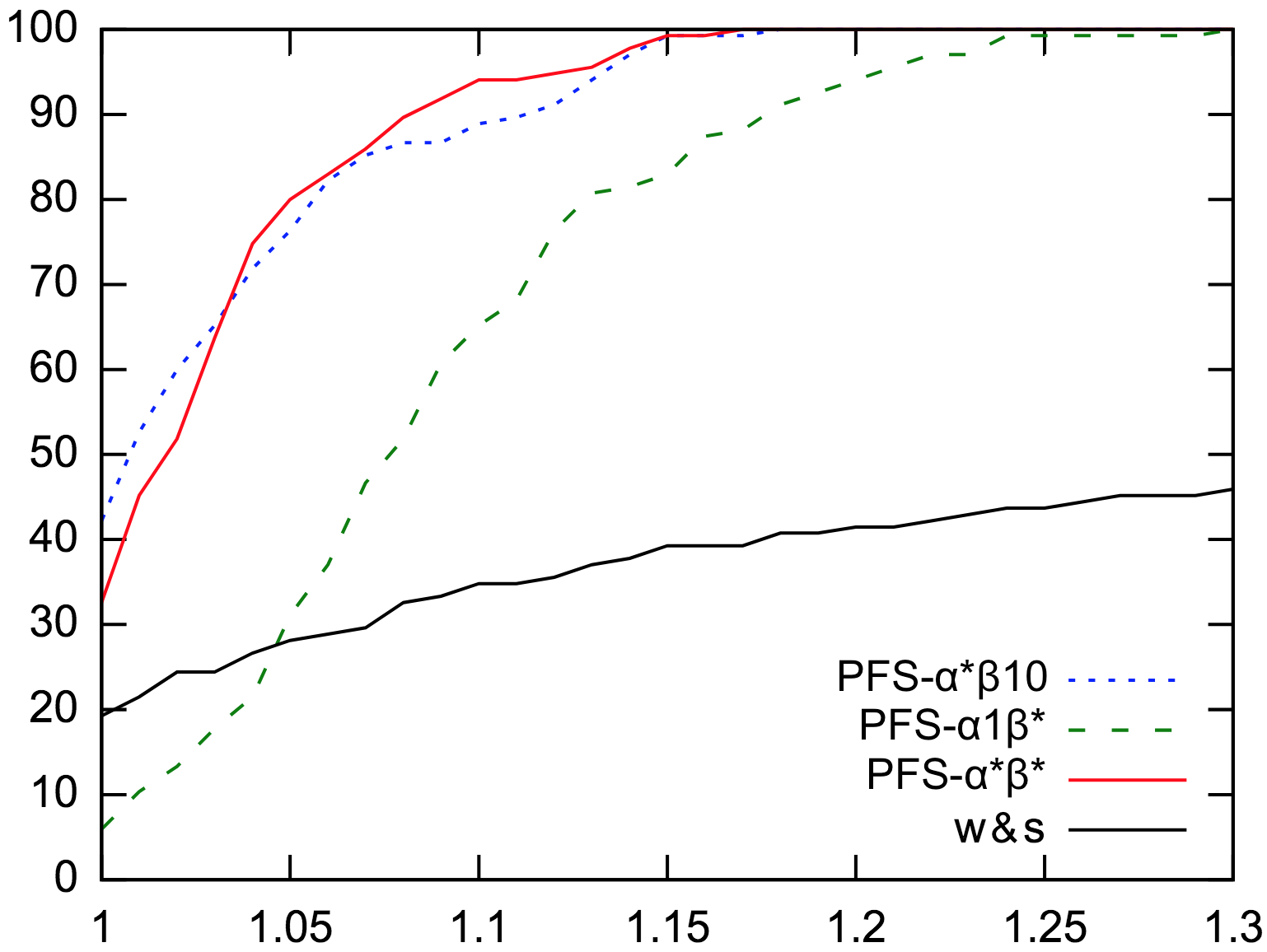}
\end{minipage}
\caption{
Performance profiles.
Left: comparison of the seven PFS instantiations and the \emph{w\&s} policy on the 15 instances of class \emph{50c+w-$i$}, using $K=20$ vehicles.
Right: comparison of PFS instantiations \emph{PFS-$\alpha$*$\beta$10}, \emph{PFS-$\alpha$1$\beta$*} and \emph{PFS-$\alpha$*$\beta$*} on the 3 classes (\emph{50c-30w-$i$}, \emph{50c-50w-$i$}, \emph{50c+w-$i$}), with $K\in \{5, 10, 20\}$ vehicles (135 instances).
}
\label{fig:performance_profiles}
\end{figure}

The performances of the seven PFS instantiations and the baseline w\&s approach are compared in Figure \ref{fig:performance_profiles} by using \emph{performance profiles}.
Performance profiles (\citeauthor{dolan2002benchmarking}, \citeyear{dolan2002benchmarking}) provide, for each considered approach, a cumulative distribution of its performance compared to other approaches. 
For a given method A, a point $(x,y)$ on A's curve means that in $(100\cdot y)\%$ of the instances, A performed at most $x$ times worse than the best method on each instance taken separately. 
A method A is strictly better than another method B if A's curve always stays above B's curve.

According to Figure \ref{fig:performance_profiles} (left), algorithms \emph{PFS-$\alpha$*$\beta$10} and \emph{PFS-$\alpha$*$\beta$*} show the best performances when tested on the 15 instances of class \emph{50c+w-$i$} with $K=20$ vehicles. 
More experiments are conducted and reported in Figure \ref{fig:performance_profiles} (right) in order to distinguish between the algorithms \emph{PFS-$\alpha$*$\beta$10}, \emph{PFS-$\alpha$1$\beta$*} and \emph{PFS-$\alpha$*$\beta$*} on all 135 instances. 
In comparison to the other approaches, algorithms \emph{PFS-$\alpha$*$\beta$10} and \emph{PFS-$\alpha$*$\beta$*} clearly obtain the best performances on average over the 135 configurations. 

\begin{figure}
\centering
\begin{minipage}[c]{.49\linewidth}
      \includegraphics[width=1.0\textwidth]{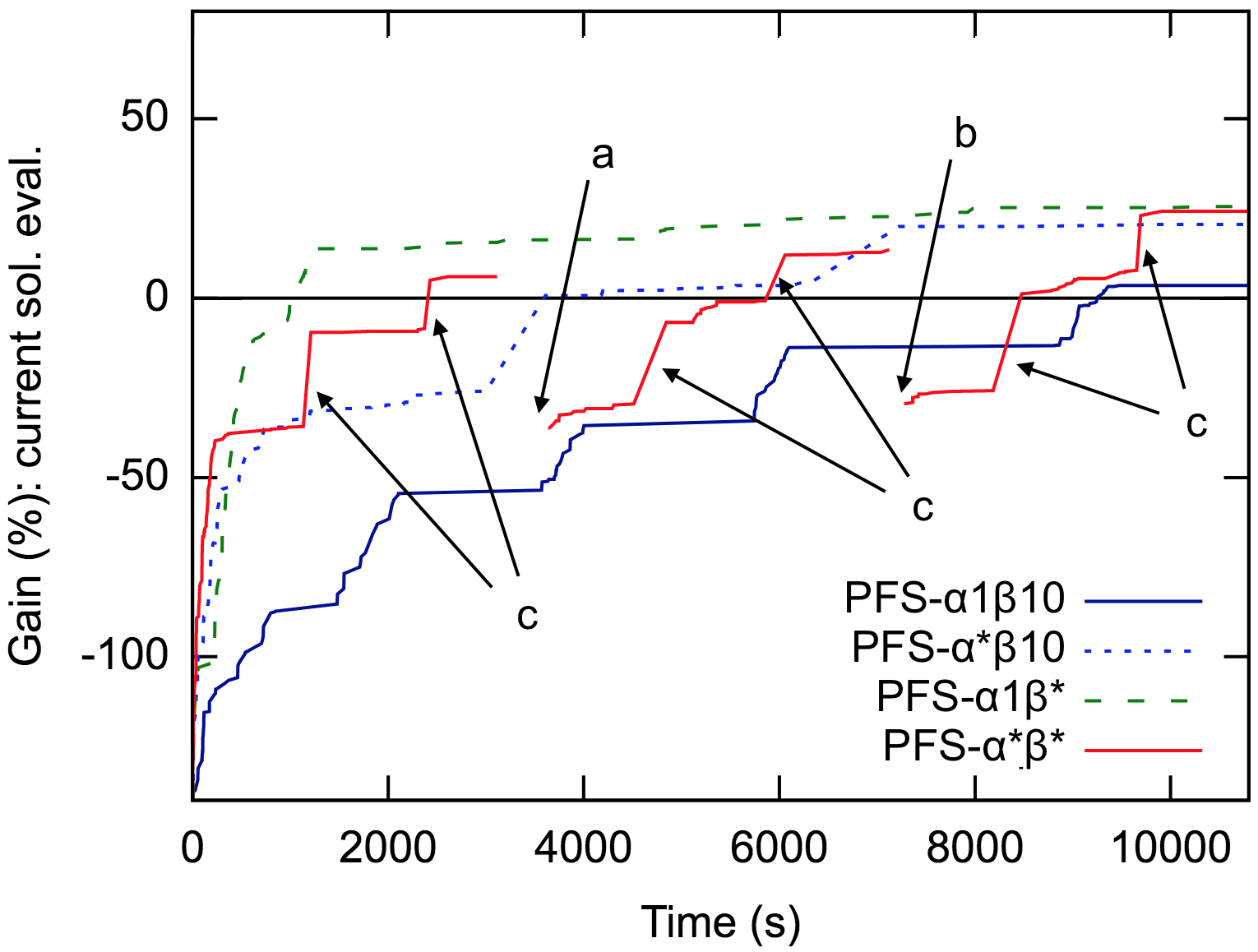}
   \end{minipage} \hfill
   \begin{minipage}[c]{.49\linewidth}
      \includegraphics[width=1.0\textwidth]{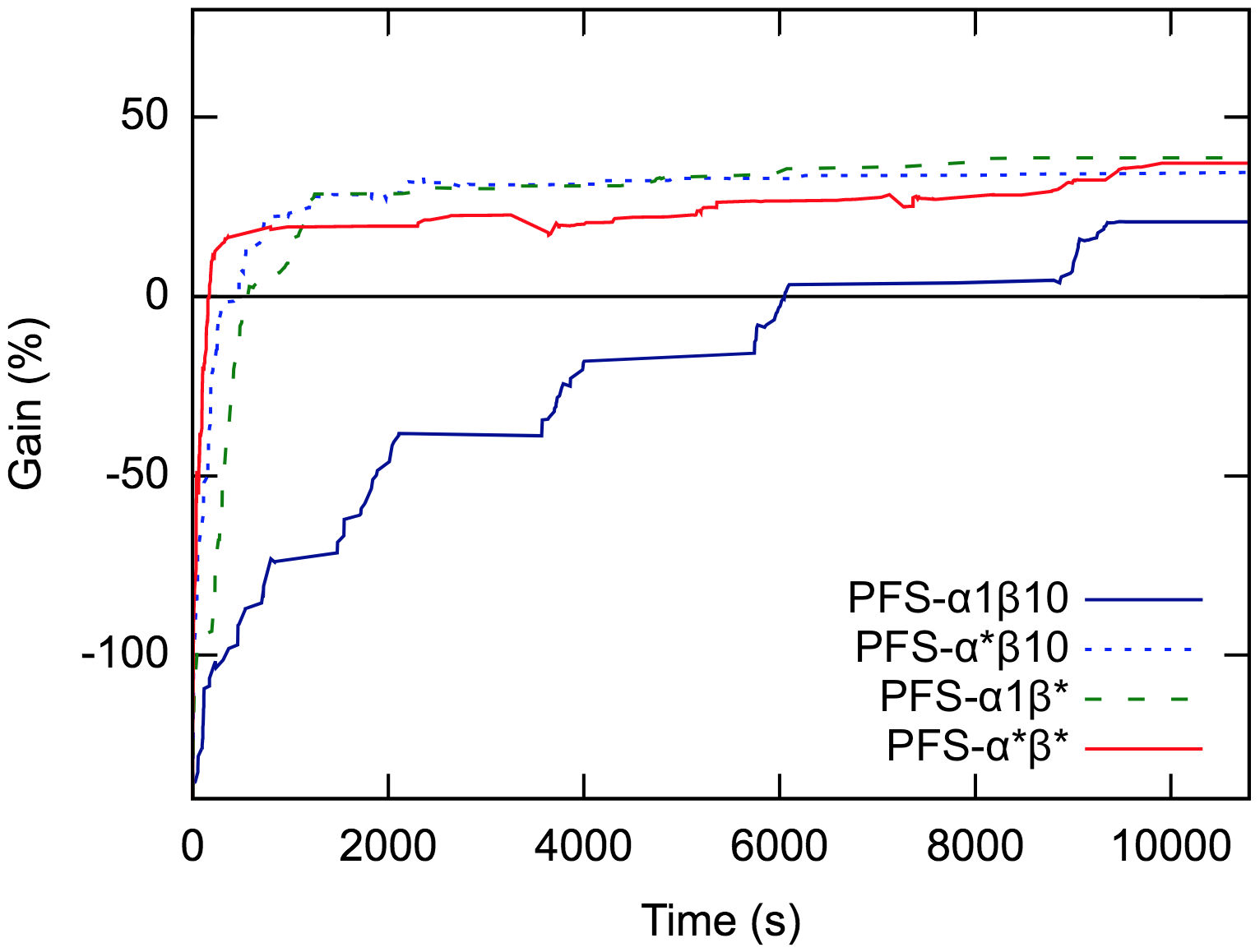}
\end{minipage}
\caption{
Evolution through time of the gain of the expected cost of the current solution with respect to the average cost of the w\&s policy, during a single execution of four PFS instantiations for instance \emph{50c-50w-1} (with $K=10$ vehicles). Left: gain evaluated under $\mathcal{R}^q$ at current scale $\alpha$. Right: gain evaluated under $\mathcal{R}^{q+}$ at scale $\alpha = 1$.
}
\label{fig:evo}
\end{figure}

Figure \ref{fig:evo} illustrates, on a single instance (\emph{50c-50w-1} with $K=10$ vehicles), the evolution through time of the gain of the expected cost of the current solution $s$, with respect to the average cost of the w\&s policy, during a single run of \emph{PFS-$\alpha$1$\beta$10}, \emph{PFS-$\alpha$*$\beta$10}, \emph{PFS-$\alpha$1$\beta$*}, and \emph{PFS-$\alpha$*$\beta$*}. 
For each incumbent solution $s$, the left part of Figure \ref{fig:evo} plots the gain of $s$ under $\mathcal{R}^q$ at its current scale $\alpha$. It corresponds to the quality of $s$ as evaluated by the LS algorithm. The right part plots the corresponding gain under $\mathcal{R}^{q+}$ at scale $\alpha = 1$.
In the left part, we clearly recognize the nine different optimization phases of \emph{PFS-$\alpha$*$\beta$*}. 
A drop in the expected cost happens whenever the current solution $s$ is converted to a higher scale factor. This happens twice during the run: from $\alpha_2=1$ to $\alpha_3=5$ (point $a$) and from $\alpha_5=1$ to $\alpha_6=5$ (point $b$). In both cases, the resulting solution becomes infeasible and the algorithm needs some time to restore feasibility.
A sudden leap happens when converting to a lower scale. This happens six times (points $c$): from $ \alpha_i=5$ to $\alpha_{i+1}=2$ and from $\alpha_{i+1}=2$ to $\alpha_{i+2}=1$, with $i\in\{0,3,6\}$. 
This is a direct consequence of the fact that rounding operations are always performed in a pessimistic way, as explained in Section \ref{sec:expe_plan}. 
Whereas the quality of $s$ under $\mathcal{R}^{q}$ at scale $\alpha$ appears to be worse than that of \emph{PFS-$\alpha$1$\beta$10} (\textit{e.g.}, at point $b$), the true gain of $s$ (evaluated under $\mathcal{R}^{q+}$, $\alpha = 1$) remains always better with \emph{PFS-$\alpha$*$\beta$*}. 

\begin{figure}[t]
\centering
\includegraphics[width=1.0\textwidth]{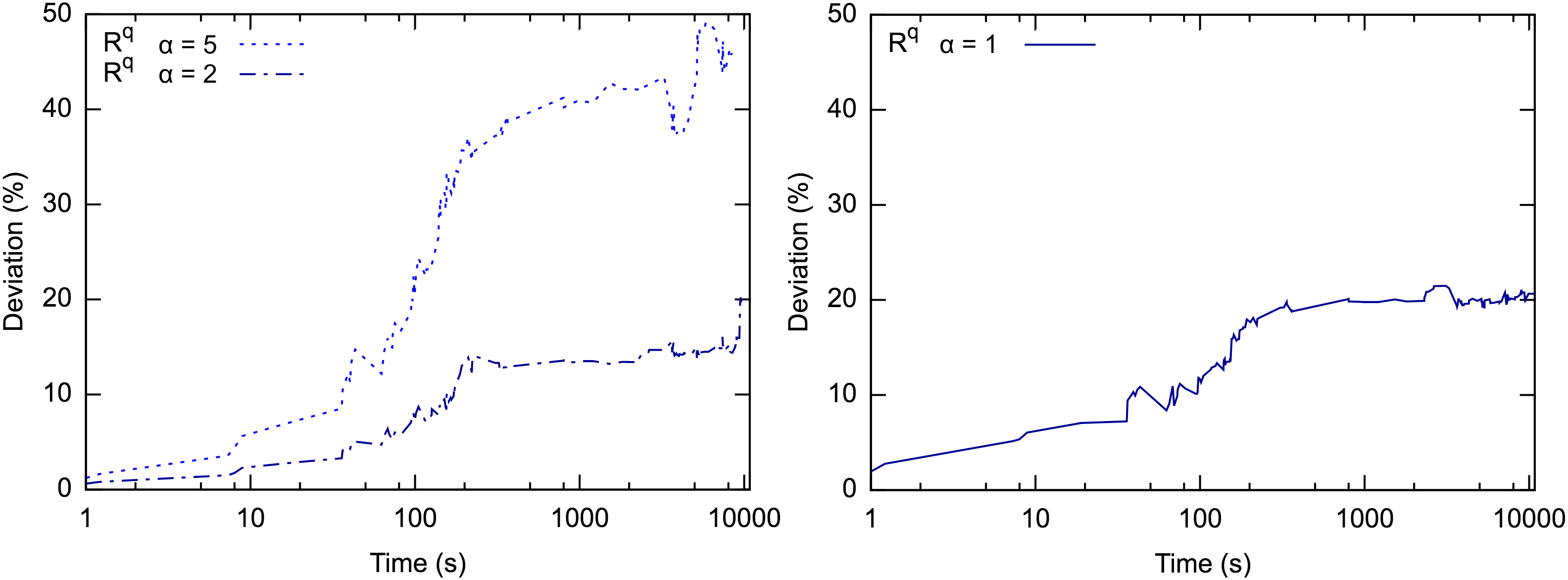}
\caption{
Scale approximation quality and impact of recourse strategies. 
For each solution $s$ encountered while running \emph{PFS-$\alpha$*$\beta$*}, on instance \emph{50c-50w-1} as displayed in Figure \ref{fig:evo}, left curves show the evolution of the gap (in \%) between costs computed with $\alpha\in\{2,5\}$ and those computed with $\alpha = 1$. On right, the gap between $\mathcal{R}^{q}$ and $\mathcal{R}^{q+}$, both with $\alpha = 1$.
}
\label{fig:dev}
\end{figure}

Finally, Figure \ref{fig:dev} compares the expected costs when varying either the scale $\alpha$ (left) or the recourse strategy (right), using the same sequences of solutions than those used for Figure \ref{fig:evo}.
On left, the evolution of the deviation (\%) between costs computed with $\alpha=1$, and $\alpha\in\{2,5\}$, under strategy $\mathcal{R}^{q}$. 
On right, the deviation between costs computed with $\mathcal{R}^{q}$ and $\mathcal{R}^{q+}$, with $\alpha=1$ in both cases. 
Scale $\alpha=2$ (left, long dashed) always provides a better approximations, closer to the one as computed under $\alpha = 1$, than scale $\alpha=5$ (left, dashed).
We also notice a significant increase in the gaps as the algorithm finds better solutions: under 100 seconds, costs computed at scale $\alpha=2$ (resp. $\alpha=5$) remain at maximum 10\% (resp. 20\%) from what would be computed under $\alpha=1$, and tend to stabilize at around 20\% (resp. 45\%) in the long term.
Similar observations can be made (Figure \ref{fig:dev}, right) regarding the gap between costs computed with $\mathcal{R}^{q}$ at $\alpha=1$ and those computed with $\mathcal{R}^{q+}$, $\alpha=1$.
Similarly, the cost difference subsequent to the recourse strategy tends to increase progressively with the quality of the solutions. 
This could be explained by the time discrepancies generated by rounding operations when a solution is scaled. Better solutions having complex, tighter schedules are then less robust to such time approximations, and more sensible to the discrepancy effects which propagate and impact on the customer time windows.

\subsection{Results on large instances \label{sec:big_instances_results}}
We now analyze how our SS-VRPTW-CR model behaves compared to the w\&s policy
, when varying both vehicle fleet size and the urgency of requests. 
We consider algorithm \emph{PFS-$\alpha$*$\beta$*} only.

\subsubsection{Influence of the number of vehicles \label{sec:big_instances_number_veh}}
Table \ref{tab:tw_x1} shows how the performance of the SS-VRPTW-CR model relative to the w\&s policy varies with the waiting locations and the number of vehicles. For 5, 10, and 20 vehicles, the average over each of the instance classes (15 instances per class) is reported. 

It shows us that the more vehicles are involved, the more important clever anticipative decisions are, and therefore the more beneficial a SS-VRPTW-CR solution is compared to the w\&s policy.
It is likely that, as conjectured in \cite{Saint-Guillain2017}, a higher number of vehicles leads to a less uniform objective function, most probably with the steepest local optima. 
Because it requires much more anticipation than when there are only five vehicles, using the SS-VRPTW-CR model instead of the w\&s policy is found to be particularly beneficial provided that there are at least 10 or 20 vehicles. With 20 vehicles, our model decreases the average number of rejected requests by 52.2\% when vehicles are allowed to wait at customer vertices (\emph{i.e.} for the class of instances \emph{50c+w-$i$}). 

On the other hand, we also observe that due to the lack of anticipative actions, the w\&s policy globally fails at tacking the advantage of a larger number of vehicles. Indeed, allowing 20 vehicles does not significantly improve the performances of the baseline policy compared to only 10 vehicles.

\begin{table}[]
\small
\centering
\begin{tabular}{lrrrrrrrrrrr}
       &  & \multicolumn{10}{c}{\emph{PFS-$\alpha$*$\beta$*}}                                                      \\ \cline{3-12} 
       &  & w\&s      &  & \multicolumn{2}{c}{50c30w} &  & \multicolumn{2}{c}{50c50w} &  & \multicolumn{2}{c}{50c+w} \\ \cline{3-3} \cline{5-6} \cline{8-9} \cline{11-12} 
       &  & \#rejects &  & \#rejects      & \%gain      &  & \#rejects      & \%gain      &  & \#rejects      & \%gain     \\
\hline
$K=5$  &  &       39.3    &  &             40.1   & -2.5      &  &             40.7   & -4.3      &  &            39.0    & 0.4      \\
$K=10$ &  &        33.9   &  &              27.1  & 18.4      &  &             25.5   & 22.9      &  &              23.6  & 29.2     \\
$K=20$ &  &        33.7   &  &              20.3  & 38.9      &  &             17.9   & 45.8      &  &           16.0     & 52.2    
\end{tabular}
\caption{Average number of rejected requests on instances \emph{50c-30w-$i$}, \emph{50c-50w-$i$}, \emph{50c+w-$i$}, \textit{w.r.t.} the number of vehicles. }
\label{tab:tw_x1}
\end{table}

\subsubsection{Influence of the time windows}
We now consider less urgent requests, by conducting the same experiments as in section \ref{sec:big_instances_number_veh} while modifying the time windows only.
Table \ref{tab:tw_x2} shows the average gain of using an SS-VRPTW-CR model when the service quality is reduced by \emph{multiplying all the original time window durations by two}. 
\begin{table}[]
\small
\centering
\begin{tabular}{lrrrrrrrrrrr}
       &  & \multicolumn{10}{c}{\emph{PFS-$\alpha$*$\beta$*}}                                                      \\ \cline{3-12} 
       &  & w\&s      &  & \multicolumn{2}{c}{50c30w} &  & \multicolumn{2}{c}{50c50w} &  & \multicolumn{2}{c}{50c+w} \\ \cline{3-3} \cline{5-6} \cline{8-9} \cline{11-12} 
       &  & \#rejects &  & \#rejects      & \%gain      &  & \#rejects      & \%gain      &  & \#rejects      & \%gain     \\
\hline
$K=5$  &  &         21.0  &  &         28.1       & -35.1      &  &          27.3      & -31.4      &  &           26.3 & -26.3      \\
$K=10$ &  &       13.8    &  &         14.8       & -11.7      &  &        13.8       & -3.5      &  &          12.9     & 3.4     \\
$K=20$ &  &       13.6    &  &         10.4      & 21.2      &  &        9.0       & 31.7      &  &           8.8 & 33.7    
\end{tabular}
\caption{Average number of rejected requests, all time window durations being doubled. }
\label{tab:tw_x2}
\end{table}
The results show that for $K=5$ vehicles, the w\&s policy always performs better. 
With $K=20$ vehicles, however, the average relative gain achieved by using the SS-VRPTW-CR model remains significant: there are 33.7\% fewer rejected requests on average for the class of instances \emph{50c+w-$i$}. 

Table \ref{tab:tw_x3} illustrates how the average gain is impacted when \emph{time windows are multiplied by three}. 
Given 20 vehicles, the SS-VRPTW-CR model still improves the w\&s policy by 14\% when vehicles are allowed to wait directly at customer vertices. 
Together with Table \ref{tab:tw_x1}, Tables \ref{tab:tw_x2} and \ref{tab:tw_x3} show that the SS-VRPTW-CR model is more beneficial when the number of vehicles is high and the time windows are small, that is, in instances that are particularly hard in terms of quality of service and thus require much more anticipation.

\begin{table}[]
\small
\centering
\begin{tabular}{lrrrrrrrrrrr}
       &  & \multicolumn{10}{c}{\emph{PFS-$\alpha$*$\beta$*}}                                                      \\ \cline{3-12} 
       &  & w\&s      &  & \multicolumn{2}{c}{50c30w} &  & \multicolumn{2}{c}{50c50w} &  & \multicolumn{2}{c}{50c+w} \\ \cline{3-3} \cline{5-6} \cline{8-9} \cline{11-12} 
       &  & \#rejects &  & \#rejects      & \%gain      &  & \#rejects      & \%gain      &  & \#rejects      & \%gain     \\
\hline
$K=5$  &  &    14.2       &  &        22.5        & -61.1      &  &        22.0        & -57.8      &  &               21.1 & -50.7      \\
$K=10$ &  &     7.4     &  &         10.9       & -60.0      &  &        9.7       & -41.7      &  &              8.8 & -26.1     \\
$K=20$ &  &      7.3     &  &         8.1       & -19.8      &  &          7.0     & -4.1      &  &       6.0       & 14.0    
\end{tabular}
\caption{Average number of rejected requests, all time window durations being tripled. }
\label{tab:tw_x3}
\end{table}

\subsubsection{Positions of the waiting locations}
From all the experiments conducted on our benchmark, it immediately appears that, no matter the operational context (number of customer vertices, vehicles) or the approximations that are used (scaling factor, waiting time multiples), \emph{allowing the vehicles to wait directly at customer vertices always leads to better results than using separated waiting vertices}. 
Unless the set of possible waiting locations is restricted, \emph{e.g.}, big vehicles cannot park anywhere in the city, placing waiting vertices in such a way that they coincide with customer vertices appears to be the best choice.

\section{Conclusions and research directions\label{sec:conclusions}}
In this paper, we consider the SS-VRPTW-CR problem previously introduced by \cite{Saint-Guillain2017}.
We extend the model with two additional recourse strategies: $\mathcal{R}^{q}$ and $\mathcal{R}^{q+}$. 
These take customer demands into account and allow the vehicles to save operational time, traveling directly between customer vertices when possible.
We show how, under these recourse strategies, the expected cost of a second-stage solution is computable in pseudo-polynomial time.

Proof of concept experiments on small and reasonably large test instances compare these anticipative models with each other and show their interest compared to a basic "wait-and-serve" policy. 
These preliminary results confirm that, although computationally more demanding, optimal first-stage solutions obtained with $\mathcal{R}^{q+}$ generally show significantly better expected behavior. 
The LS algorithm presented in \cite{Saint-Guillain2017} produces near-optimal solutions on small instances.

In this paper, we also introduce PFS, a meta-heuristic particularly suitable for our problem when coupled with the LS algorithm. 
More generally, PFS is applicable to any problem in which: 
\textit{a)} the objective function is particularly complex to compute but depends on the accuracy of the data
and \textit{b)} the size of the solution space can be controlled by varying the granularity of the operational decisions.
We show that PFS allows to efficiently tackle larger problems for which an exact approach is not possible. 

We show that SS-VRPTW-CR recourse strategies provide significant benefits compared to a basic, non-anticipative but yet realistic policy.
Results for a variety of large instances show that the benefit of using the SS-VRPTW-CR increases with the number of vehicles involved and the urgency of the requests.
Finally, all our experiments indicate that allowing the vehicles to wait directly at potential customer vertices, when applicable, leads to better expected results than using separated relocation vertices. 

\subsection*{Future work and research avenues}

\paragraph{On solution methods.}
An adaptive version of PFS, therefore improving the algorithm by making dynamic the decision about changing the scale factor $\alpha$ or the domain reduction factor $\beta$, could be designed.
Exact optimal methods should also be investigated. However, the black box nature of the evaluation function $\mathcal{Q^R}$ makes classical (stochastic) integer programming approaches (\textit{e.g.} branch-and-cut, L-shaped method, \textit{etc.}) unsuitable for the SS-VRPTW-CR unless efficient valid inequalities that are active at fractional solutions can be devised (such as those proposed by \citeauthor{Hjorring1999}, \citeyear{Hjorring1999}, for the SS-VRP-D). 
Amongst other possible candidates for solving this problem, we could consider set-partitioning methods such as column generation, which are becoming commonly used for stochastic VRPs. 
Approximate Dynamic Programming (ADP, \cite{Powell2009}) is also widely used to solve routing problems in presence of uncertainty. Combined with scaling techniques, ADP is likely to provide interesting results.

\paragraph{On scaling techniques.}
We have shown through experiments that the computational complexity of the objective function is an issue that can be successfully addressed by scaling down problem instances. 
However, the scale is only performed in terms of temporal data, decreasing the accuracy of the time horizon. 
It may also be valuable to consider a reduced, clustered set of potential requests, which would also allow us to significantly reduce  computational effort when evaluating a first-stage solution.

\paragraph{Further application to online optimization.}
As already pointed out by \cite{Saint-Guillain2017}, another potential application of the SS-VRPTW-CR is to online optimization problems such as the Dynamic and Stochastic VRPTW (DS-VRPTW). 
Most of the approaches that have been proposed in order to solve the DS-VRPTW rely on reoptimization. 
However, because perfect online reoptimization is intractable, heuristic methods are often preferred. 
Approaches based on sampling, such as Sample Average Approximation (\citeauthor{ahmed2002sample}, \citeyear{ahmed2002sample}), are very common and consist in restricting the set of scenario to a random subset. 
Because the computed costs depend on the quality and size of the subset of scenarios, they do not provide any guarantee.
Thanks to recourse strategies, the expected cost of a first-stage SS-VRPTW-CR solution provides an upper bound on the expected cost under perfect reoptimization, as it also enforce the nonanticipativity constraints (see \citeauthor{Saint-Guillain2015}, \citeyear{Saint-Guillain2015}, for a description of these constraints).
The SS-VRPTW-CR can therefore be exploited when solving the DS-VRTPW.

\paragraph{Towards better recourse strategies.}
The expected cost of a first-stage solution obviously depends on how the recourse strategy fits the operational problem.
Improving these strategies may tremendously improve the quality of the upper bound they provide to exact reoptimization.
The recourse strategies presented in this paper are of limited operational complexity, yet their computational complexity is already very expensive. 
One potential improvement which would limit the increase in computational requirements would be to rethink the way in which the potential requests are assigned to waiting locations, \emph{e.g.} by taking their probabilities and demands into account. 
Another direction would be to think about better, more intelligent, vehicle operations.
However, an important question remains: how intelligent could a recourse strategy be such that its expected cost stays efficiently computable?

\newpage
\begin{appendices}

\section{Stochastic Integer Programming formulation of the SS-VRPTW-CR \label{sec:sip}}

The problem stated by (\ref{eq:ss-vrptw-general})-(\ref{eq:ss-vrptw-general-2}) refers to a nonlinear stochastic integer program with recourse, which can be modeled as the following simple extended three-index vehicle flow formulation:
\begin{align}
& \underset{x,\tau}{\text{Minimize}} \hspace{1em}  \mathcal{Q}^\mathcal{R}(x,\tau) \label{objective-function} \\
&\text{subject to}& \notag\\
& \hspace{1em} \sum_{j \in W_0} x_{ijk} = \sum_{j \in W_0} x_{jik} = y_{ik}  && \forall ~ i \in W_0, ~k \in [1,K] \label{degree-2} \\
& \hspace{1em} \sum_{k \in [1,K]} y_{0k} \le K && \label{depot-visited} \\
& \hspace{1em} \sum_{k \in [1,K]} y_{ik} \le 1 &&\forall ~ i \in W \label{customer-visited} \\
& \hspace{1em} \sum_{\substack{i \in S \\ j \in W\setminus S}} x_{ijk}  \ge y_{vk} && \forall ~ S \subseteq W, ~ v \in S, ~k \in [1,K] \label{subtour-elim} \\
& \hspace{1em} \sum_{l \in H} \tau_{ilk} = y_{ik}   && \forall ~i \in W, ~ k \in [1,K] \label{tau-variables-one}  \\
& \hspace{1em} \sum_{\substack{i \in W_0 \\ j \in W_0 }} x_{ijk} ~ d_{i,j}  + \sum_{\substack{i \in W \\ l \in H}} \tau_{ilk} ~ l +1 \le h   &&\forall ~ k \in [1,K]  \label{route_length}  \\
& \hspace{1em} y_{ik} \in \{0,1\}  && \forall ~ i \in W_0,  ~ k \in [1,K] \label{y-variables} \\
& \hspace{1em} x_{ijk} \in \{0,1\}  && \forall ~ i,j \in W_0 : ~i \ne j, ~ k \in [1,K]  \label{x-variables} \\
& \hspace{1em}  \tau_{ilk} ~ \in \{0,1\}  && \forall ~ i \in W,  ~ l \in H, ~k \in [1,K]  \label{tau-variables}  
\end{align}
Our formulation uses the following binary decision variables:
\begin{itemize}
\item $y_{ik}~$ equals $1$ iff vertex $i \in W_0$ is visited by vehicle (or route) $k \in [1,K]$;
\item $x_{ijk}~$ equals $1$ iff the arc $(i,j)\in W_0^2$ is part of route $k \in K$; 
\item $\tau_{ilk}~~$ equals $1$ iff vehicle $k$ waits for $1 \le l \le h$ time units at vertex $i$.
\end{itemize}
Whereas variables $y_{ik}$ are only of modeling purposes, yet  $x_{ijk}$ and $\tau_{ilk}$ variables solely define a SS-VRPTW-CR first stage solution.
Constraints (\ref{degree-2}) to (\ref{subtour-elim}) together with (\ref{x-variables}) define the feasible space of the asymmetric Team Orienteering Problem (\citeauthor{Chao1996}, \citeyear{Chao1996}). 
In particular, constraint (\ref{depot-visited}) limits the number of available vehicles.
Constraints (\ref{customer-visited}) ensure that each waiting vertex is visited at most once. 
Subtour elimination constraints (\ref{subtour-elim}) forbid routes that do not include the depot. 
Constraint (\ref{tau-variables-one}) ensures that exactly one waiting time $1 \le l \le h$ is selected for each visited vertex. 
Finally, constraint (\ref{route_length}) states that the total duration of each route, starting at time unit 1, cannot exceed $h$.



\section{Expected cost of second-stage solutions under $\mathcal{R}^q$ \label{sec:R_Q_expectation}}

In this section we explain how the expected cost of second stage solutions, provided a first stage solution $(x, \tau)$ to the SS-VRPTW-CR, can be efficiently computed in the case of recourse strategy $\mathcal{R}^{q}$.
As a reminder, $\mathcal{R}^{q}$ generalizes strategy $\mathcal{R}^{\infty}$, introduced in \cite{Saint-Guillain2017}, by considering vehicle capacity constraints.

Recall also that, once the request ordering and assignment phase finished, we end up with a partition $\{ \pi_\bot, \pi_1, ..., \pi_K \}$ of $R$,
where $\pi_k$ is the ordered sequence of potential requests assigned to the waiting vertices visited by vehicle $k$, and $\pi_\bot$ is the set of unassigned requests (such that $\w(r)=\bot$).
We note $\pi_w$, the set of requests assigned to a waiting vertex $w \in W^x$.
We note $\fst(\pi_w)$ and $\fst(\pi_k)$, the first requests of $\pi_w$ and $\pi_k$, respectively, according to the order $<_R$. For each request $r \in \pi_k$ such that $r \neq \fst(\pi_k)$, we note $\prv(r)$, the request of $\pi_k$ that immediately precedes $r$ according to the order $<_R$.
Table \ref{table:notations_2} summarizes the main notations introduced in this section.
Remember that they are all specific to a first-stage solution $(x, \tau)$.
\begin{table}[h]
\caption{\label{table:notations_2}
Notations summary: material for recourse strategies.}
\begin{tabular}{lclll}
\hline
$\bot$				 &~~~& The null vertex: $\forall r \in R, \w(r) = \bot \Leftrightarrow r$ is unassigned \\
$\w(r)$					&& Waiting vertex of $W^x$ to which $r \in R$ is assigned \\
$\pi_k$					&& Potential request assigned to vehicle $k$: ~~$\pi_k = \{ r \in R : \w(r) \in x_k \}$		\\
$\pi_w$					&& Potential request assigned to waiting location $w \in W^x$: ~~$\pi_w = \{ r \in R : \w(r) = w \}$\\
$\fst(\pi_w)$			&& Smallest request of $\pi_w$ according to $<_R$. 		\\
$\fst(\pi_k)$			&& Smallest request of $\pi_k$ according to $<_R$. 		\\
$\prv(r)$ 				&& Request of $\pi_k$ which immediately precedes $r$ according to $<_R$, if any 	\\
$t^\text{min}_{r,w}$	&& Min. time from which a vehicle can handle request $r \in R$ from $w \in W^x$ \\
$t^\text{max}_{r,w}$	&& Max. time from which a vehicle can handle request $r \in R$ from $w \in W^x$ \\
\hline
\end{tabular}
\end{table}
In the case of strategy $\mathcal{R}^{q}$, $t^\text{min}_{r,w}$ and $t^\text{max}_{r,w}$ are computed according to the definition provided in \cite{Saint-Guillain2017} for strategy $\mathcal{R}^{\infty}$. Hence,  
$t^\text{min}_{r, w}= \max\{\underline{on}(w), ~ \Gamma_{r}, ~ e_{r} - d_{w,{r}}\}$ 
and 
$t^\text{max}_{r,w} = \min \{ l_{r} - d_{w,r} , ~ \overline{on}(w) - d_{w,r} - s_{r} - d_{r, w} \}$.

We assume that request probabilities to be independent of each other; {\em i.e.}, for any couple of requests $r, r' \in R$, the probability $p_{r \wedge r'}$ that both requests will appear is given by $p_{r \wedge r'} = p_r \cdot p_{r'}$.

$\mathcal{Q}^{\mathcal{R}^q}(x,\tau)$ is equal to the expected number of rejected requests, which in turn is equal to the expected number of requests that are found to appear minus the expected number of accepted requests. Under the independence hypothesis, the expected number of revealed requests is given by the sum of all request probabilities, whereas the expected number of accepted requests is equal to the cumulative sum, for every request $r$, of the probability that it belongs to $A^h$, {\em i.e.}, 
\begin{equation}
\mathcal{Q}^{\mathcal{R}^q}(x,\tau) = \sum_{r \in R} p_r - \sum_{r \in R} \text{Pr}\{ r \in A^h \} 
= \sum_{r \in R} \big( p_r - \text{Pr}\{ r \in A^h \} \big) \label{eq:expectation_infinite}
\end{equation}
The probability $\text{Pr}\{r\in A^h\}$ is computed by considering every feasible time $t\in [ t^\text{min}_{r, w},t^\text{max}_{r, w}]$ and every possible load configuration $q\in [0,Q-q_r]$ that satisfies $r$:
\begin{align}
\text{Pr}\{ r \in A^h \} &= \sum_{t = t^\text{min}_{r, w}}^{t^\text{max}_{r, w}} \sum_{q=0}^{Q-q_r} g_1(r, t, q) \label{eq:req-satisfiable_capacity} .
\end{align}
$g_1(r,t,q)$ is the probability that $r$ has appeared and that vehicle $k$ leaves $\w(r)$ at time $t$ with load $q$ to serve $r$, {\em i.e.}, 
\begin{align}
g_1(r,t, q) \equiv \text{Pr}\{ &r \text{ appeared}, \departureTime(r)=t \text{ and }\load(k,t)=q\} \notag 
\end{align}
where $\load(k,t)$ is the load of vehicle $k\in [1,K]$ at time $t\in H$, and $\departureTime(r) = \max\{{\it available}(r), t^\text{min}_{r,\w(r)}\}$ is the time at which it actually leaves the waiting vertex $\w(r)$ in order to serve $r$ (the vehicle may have to wait if ${\it available}(r)$ is smaller than the earliest time for leaving $\w(r)$ to serve $r$).

\paragraph{Computation of probability $g_1(r,t,q)$}
Recall that $\pi_k$ is the set of potential requests on route $k \in [1,K]$, ordered by $<_R$.
The base case for computing $g_1$ is concerned with the very first potential request on the entire route, $r=\fst(\pi_k)$, which must be considered as soon as vehicle $k$ arrives at $w = \w(r)$, that is, at time $\underline{on}(w)$, except if 
$\underline{on}(w) < t^\text{min}_{r, w}$:
\begin{align}
\mbox{if } &r = \fst(\pi_k) \mbox{ then }\notag\\
&g_1(r,t,q) = 
\begin{cases}
p_{r} & \text{if } ~~ t = \max\{ \underline{on}(w), t^\text{min}_{r, w} \} ~~\wedge~~ q = 0 \\ 
0 & \text{otherwise}.
\end{cases} \label{eq:g1_basic_base}
\end{align}
For any $q \ge 1$, $g_1(r,t,q)$ is equal to zero as vehicle $k$ necessarily carries an empty load when considering the first request $r$.

The more general case of a request $r$ which is not the first request of a waiting vertex $w \in W^x$, ({\em i.e.}, $w\ne \fst(\pi_w)$), depends on the time and load configuration at which vehicle $k$ is available for $r$,
Although ${\it available}(r)$ and $\load(k,t)$ are both deterministic when we know the set $A^{\Gamma_{\prv(r)}}$ of previously accepted requests, this is not true anymore when computing probability $g_1(r,t,q)$. 
As a consequence, $g_1(r,t,q)$ depends on the probability $f(r,t,q)$ that vehicle $k$ is available for $r$ at time $t$ with load $q$:
$$f(r,t,q) \equiv \text{Pr}\{ {\it finishToServe}(\prv(r))=t \text{ and }\load(k,t)=q\}.$$
Note that for any such request $r \in R: r \ne \fst(\pi_{\w(r)})$, the time ${\it finishToServe}(\prv(r))$ is equivalent to ${\it available}(r)$. On the contrary, this is not the case for a request that is the first of its waiting vertex.
The computation of $f$ is detailed below.
Given this probability $f$, the general case for computing $g_1$ is:
\begin{align}
\mbox{if } &r \neq \fst(\pi_{\w(r)}) \mbox{ then }\notag\\ 
&g_1(r,t, q) = \begin{cases}
p_{r} \cdot f(r,t, q) & \text{if } ~~ t > t^\text{min}_{r,\w(r)} \\
p_{r} \cdot \sum_{t' = \underline{on}(\w(r))}^{t^\text{min}_{r,\w(r)}} 
f(r,t' , q) ~~~ & \text{if } ~~ t = t^\text{min}_{r,\w(r)} \\
0 & \text{otherwise} 
\end{cases} \label{eq:g1_R_capa}
\end{align}
Indeed, if $t > t^\text{min}_{r,\w(r)}$, then vehicle $k$ leaves $\w(r)$ to serve $r$ as soon as it becomes available. 
If $t < t^\text{min}_{r,\w(r)}$, the probability that vehicle $k$ leaves $\w(r)$ at time $t$ is null since $t^\text{min}_{r,\w(r)}$ is the earliest time for serving $r$ from $\w(r)$. 
Finally, at time $t = t^\text{min}_{r,\w(r)}$, we must consider the possibility that vehicle $k$ has been waiting to serve $r$ since an earlier time $\underline{on}(\w(r)) \le t' < t^\text{min}_{r,\w(r)}$. 
In this case, the probability that vehicle $k$ leaves $\w(r)$ to serve $r$ at time $t$ is $p_{r}$ times the probability that vehicle $k$ has actually been available from a time $\underline{on}(\w(r)) \le t' \le t^\text{min}_{r,\w(r)}$.

We complete the computation of $g_1$ with the particular case of a request $r$ which is not the first of the route ({\em i.e.}, $r \ne \fst(\pi_k)$) but is the first assigned to the waiting vertex associated with $r$ ({\em i.e.}, $r = \fst(\pi_{\w(r)})$). 
As the arrival time on $\w(r)$ is fixed by the first-stage solution, $\departureTime(r)$ is necessarily $\max( \underline{on}(\w(r)), t^\text{min}_{r, \w(r)})$. In particular, time ${\it finishToServe}(\prv(r))$ is no longer equivalent to ${\it available}(r)$.
Unlike $\departureTime(r)$, $\load(k,t)$ is not deterministic but rather depends on what happened previously. 
More precisely, $\load(k,t)$ depends on the load carried by vehicle $k$ when it has finished serving $\prv(r)$ at the previous waiting location $\w(\prv(r))$.
For every first request of a waiting vertex, but not the first of the route, we then have:
\begin{align}
\mbox{if } &r=\fst(\pi_{\w(r)}) \mbox{ and }r \neq \fst(\pi_k) \mbox{ then }\notag\\ 
&g_1(r,t,q) = \begin{cases}
p_{r} \cdot \sum_{t' = \underline{on}(\w(\prv(r))}^{ \overline{on}(\w(\prv(r)))} f(r,t',q) , & \text{if } ~~ t = \max( \underline{on}(\w(r)), t^\text{min}_{r, \w(r)} ) \\ 
0 & \text{otherwise}
\end{cases}, \label{eq:g1_basic_base_2}
\end{align}
where we see that all possible time units for vehicle $k$ to serve $\prv(r)$ belong to $\big[\underline{on}(\w(\prv(r))), \overline{on}(\w(\prv(r)))\big]$.

\paragraph{Computation of probability $f(r,t,q)$}
Let us now define how to compute $f(r,t,q)$, the probability that vehicle $k$ becomes available for $r$ at time $t$ with load $q$.
This depends on what happened to the previous request $r^- = \prv(r)$. 
We have to consider three cases: (a) $r^-$ appeared and was satisfied, (b) $r^-$ appeared but was rejected, and (c) $r^-$ did not appear.
Let us introduce our last probability $g_2(r,t,q)$, which is the probability that a request $r$ did not appear and is discarded at time $t$ while the associated vehicle carries load $q$. 
We note $\text{discardedTime}(r) = \max\{{\it available}(r),\Gamma_r\}$, the time at which the vehicle becomes available for $r$ whereas $r$ does not appear:
\begin{align}
g_2(r,t,q) &\equiv \text{Pr} \{ r \text{ did not appear}, \discardedTime(r)=t \text{ and }\load(k,t)=q\}. \notag
\end{align}
The computation of $g_2$ is detailed below. 
Given $g_2$, we compute $f$ as follows:
\begin{align}
f(r,t,q) = ~ & g_1(r^-, t - S_{r^-}, q - q_{r^-}) \cdot \delta(r^-, t - S_{r^-}, q-q_{r^-}) \notag \\
& + ~ g_1({r^-}, t , q) \cdot \big( 1-\delta({r^-}, t,q) \big) ~ + ~ g_2({r^-},t,q) . \label{eq:f_capa} 
\end{align}
where the indicator function $\delta(r, t, q)$ returns $1$ if and only if request $r$ is satisfiable from vertex $\w(r)$ at time $t$ with load $q$; {\em i.e.}, $\delta(r, t, q) = 1$ if $t \le t^\text{max}_{r,\w(r)}$ and $q+q_r\leq Q$, whereas $\delta(r, t, q) = 0$ otherwise.
The first term in the summation of the right hand side of equation (\ref{eq:f_capa}) gives the probability that request $r^-$ actually appeared and was satisfied (case {\it a}). 
In such a case, $\departureTime(r^-)$ must be the current time $t$ minus the delay $S_{r^-}$ needed to serve $r^-$.
The second and third terms of equation (\ref{eq:f_capa}) add the probability that the vehicle was available at time $t$ but that request $r^-$ did not consume any operational time.
There are only two possible reasons for that: either $r^-$ actually appeared but was not satisfiable (case {\it b}, corresponding to the second term) or $r^-$ did not appear at all (case {\it c}, corresponding to the third term).
Note that $f(r,t,q)$ must be defined only when $r$ is not the first potential request of a waiting location.

\paragraph{Computation of probability $g_2(r,t,q)$} 
This probability is computed recursively, as for $g_1$.
For the very first request of the route of vehicle $k$, we have:
\begin{align}
\mbox{if } &r = \fst(\pi_k) \mbox{ then }\notag\\ 
&g_2(r,t,q) = \begin{cases}
1 - p_{r}, & \text{if } ~~ t = \max( \underline{on}(\w(r)), \Gamma_{r}) ~~\wedge~~ q = 0 \\ 
0 & \text{otherwise}.
\end{cases} \label{eq:g2_basic_base}
\end{align}
The general case of a request which is not the first of its waiting vertex is quite similar to the one of function $g_1$. 
We just consider the probability $1- p_{r}$ that $r$ is found not to appear and replace $t^\text{min}_{r,\w(r)}$ by the reveal time $\Gamma_{r}$:\\
\begin{align}
\mbox{if } &r \neq \fst(\pi_{\w(r)}) \mbox{ then }\notag\\ 
&g_2(r,t, q) = \begin{cases}
(1 - p_{r}) \cdot f(\prv(r),t, q) & \text{if } ~~ t > \max( \underline{on}(\w(r)), \Gamma_r ) \\
(1 - p_{r}) \cdot \sum_{t' = \underline{on}(\w(r))}^{\max( \underline{on}(\w(r)), \Gamma_r ) } 
f(\prv(r),t' , q) ~~~ & \text{if } ~~ t = \max( \underline{on}(\w(r)), \Gamma_r ) \\
0 & \text{otherwise}. 
\end{cases} \label{eq:g2_R_capa}
\end{align}
Finally, for the first request of a waiting location that is not the first of its route, we have:
\begin{align}
\mbox{if } &r = \fst(\pi_{\w(r)})\mbox{ and }r\neq \fst(\pi_k) \mbox{ then }\notag\\ 
&g_2(r,t,q) = \begin{cases}
(1 - p_{r}) \cdot \sum_{t' = \underline{on}(\w(\prv(r)))}^{ \overline{on}(\w(\prv(r)))} f(\prv(r),t',q), & \text{if } ~~ t = \max( \underline{on}(\w(r)), \Gamma_{r} ) \\ 
0 & \text{otherwise}.
\end{cases} \label{eq:g2_basic_base_2}
\end{align}

\subsubsection*{Computational complexity.}
The complexity of computing $\mathcal{Q}^{\mathcal{R}^q}(x,\tau)$ is equivalent to that of filling up $K$ matrices of size $|\pi_k| \times h \times Q$ containing all the $g_1(r,t,q)$ probabilities. 
In particular, once the probabilities in cells $(\prv(r),1\cdots t, 1\cdots q)$ are known, the cell $(r,t,q)$ such that $r \ne \fst(\pi_w)$ can be computed in $\mathcal{O}(h)$ according to equation (\ref{eq:g1_R_capa}). 
Given $n$ customer vertices and a time horizon of length $h$, there are at most $|R| = nh \ge \sum_{k=1}^K |\pi_k|$ potential requests in total, leading to an overall worst case complexity of $\mathcal{O}(nh^2Q)$.

\subsubsection*{Incremental computation.}
Since we are interested in computing $\text{Pr}\{ r \in A^h \}$ for each request $r$ separately, by following the definition of $g_1$ and $f$, the probability of satisfying $r$ only depends on the $g_1$ and $g_2$ probabilities associated with $\prv(r)$. 
As a consequence, two similar first-stage solutions are likely to share equivalent subsets of probabilities. 
This is of particular interest when considering LS-based methods generating sequences of (first-stage) solutions, where each new solution is usually quite similar to the previous one. 
In fact, for every two similar solutions, subsets of equivalent probabilities can easily be deduced, hence allowing an incremental update of the expected cost. This does not change the time complexity, as in the worst case ({\em i.e.}, when the first waiting vertex of each sequence in $x$ has been changed), all probabilities must be recomputed. However, this greatly improves the efficiency in practice. 

\subsection{Relation with SS-VRP-C \label{sec:R_Q_ss-vrp-c}}
As presented in section \ref{sec:Introduction}, the SS-VRP-C differs by having stochastic binary demands (which represent the random customer presence) and no time window. 
In this case, the goal is to minimize the expected distance traveled, provided that when a vehicle reaches its maximal capacity, it unloads by making a round trip to the depot.
In order to compute the expected length of a first-stage solution that visits all customers, a key point is to compute the probability distribution of the vehicle's current load when reaching a customer. 
In fact, this is directly related to the probability that the vehicle makes a round trip to the depot to unload, which is denoted by the function “$f(m,r)$” in \cite{Bertsimas1992}.

Here we highlight the relation between SS-VRPTW-CR and SS-VRP-C by showing how \citeauthor{Bertsimas1992}'s “$f(m,r)$” equation can be derived from equation (\ref{eq:f_capa}) when time windows are not taken into account.

Since there is no time window consideration, we can state that $\Gamma_r = t^\text{min}_{r,w} = 1$ and $t^\text{max}_{r,w} = +\infty$ for any request $r$. Also, each demand $q_r$ is equal to 1.
Consequently, the $\delta$-function used in the computation of the $f$ probabilities depends only on $q$ and is equal to 1 if $q \le Q$. Therefore, the $f$ probabilities are defined by:
\begin{align}
f(r,t,q) = g_1(r^-, t - S_{r^-}, q-1) ~ + ~ g_2(r^-, t ,q) \notag
\end{align}
with $r^-=\prv(r)$.
Now let $f'(r,q) = \sum_{t \in H} f(r,t,q)$.
As $f(r,t,q)$ is the probability that the vehicle is available for $r$ at time $t$ with load $q$, $f'(r,q)$ is the probability that the vehicle is available for $r$ with load $q$ during the day.
It is also true that $f'(r,q)$ gives the probability that exactly $q$ requests among the $r_1, ..., r^-$ potential ones actually appear (with a unit demand).
We have:
\begin{align}
f'(r,q) = \sum_{t \in H} f(r,t,q) = \sum_{t \in H} g_1(r^-, t - S_{r^-}, q-1) ~ + ~ \sum_{t \in H} g_2(r^-, t ,q) \notag 
\end{align}
As we are interested in $f'(r,q)$, not the travel distance, we can assume that all potential requests are assigned to the same waiting vertex.
Then either $r = \fst(\pi_k)$ or $r \ne \fst(\pi_{\w(r)})$. 
If $r = \fst(\pi_k)$ we naturally obtain:
\begin{align}
f'(r,q) &= \sum_{t \in H} p_{r} \cdot f(r, t , q-1) + \sum_{t \in H} (1 - p_{r}) \cdot f(r, t, q) \notag \\
&= \begin{cases}
p_r + (1 - p_r) = 1, & \text{if } ~~ q = 0 \\ 
0, & \text{otherwise}.
\end{cases} \notag
\end{align}
If $r \ne \fst(\pi_{\w(r)})$, since we always have $t \ge t^\text{min}_{r,w}$, we have:
\begin{align}
f'(r,q) &= \sum_{t \in H} p_{r} \cdot f(r, t , q-1) + \sum_{t \in H} (1 - p_{r}) \cdot f(r, t, q) \notag \\
&= p_{r} \cdot f'(r, q-1) ~ + ~ (1 - p_{r}) \cdot f'(r, q). \notag
\end{align}
We directly see that the definition of $f'(r,q)$ is exactly the same as the corresponding function “$f(m,r)$” described in \cite{Bertsimas1992} for the SS-VRP-CD with unit demands, that is, the SS-VRP-C.

\section{Computing $g_1^v(r,t,q)$ under strategy $\mathcal{R}^{q+}$ \label{g1_Rplus}}
This section develops equation (\ref{eq:req-satisfiable-R_Q_plus}) introduced in Section \ref{sec:expected_cost_RQ_plus} for computing the expected cost of second stage solutions under recourse strategy $\mathcal{R}^{q+}$, provided a first stage solution $(x, \tau)$.
Since the method is (more complex but) similar to the one developed in the case of strategy $\mathcal{R}^{q}$ (see Section \ref{sec:r_q}), for a better understanding we recommend to read through Appendix \ref{sec:R_Q_expectation} before considering this section.

We are interested in the computation of $g_1^v(r,t,q)$, the probability that request $r$ appeared at a time $t' \le t$ and that, if it is accepted, the vehicle serves it by leaving vertex $v \in W \cup C$ at time $t$ whilst carrying a load of $q$.
Let the following additional random functions:
\begin{align}
h^w(r,t,q) \equiv \text{Pr}\{ 	& \text{the vehicle gets rid of request $r$ at time $t$ with a load of $q$} \notag \\ 
							& \text{and at waiting location $w$} \} \notag \\
h^{r'}(r,t,q) \equiv \text{Pr}\{ 	& \text{the vehicle gets rid of request $r$ at time $t$ with a load of $q$} \notag \\ 
							& \text{and at location $c_{r'}$} \} \notag
\end{align}
and:
\begin{align}
g_2^w(r, t, q) \equiv \text{Pr}\{ 	& \text{request $r$ did not appear and the vehicle discards it at time $t$} \notag \\ 
							& \text{with a load of $q$ while being at waiting location $w$} \} \notag \\
g_2^{r'}(r, t, q) \equiv \text{Pr}\{ 	& \text{request $r$ did not appear and the vehicle discards it at time $t$} \notag \\ 
							& \text{with a load of $q$ while being at location $c_{r'}$} \}. \notag
\end{align}
For the very first request $r_1^k = \fst(\pi_k)$ of the route, trivially the current load $q$ of the vehicle must be zero, and it seems normal for the waiting location $w = \w(r_1^k)$ to be the only possible location from which the vehicle can be available to handling $r_1^k$ if the request appears, or to discard it if it doesn't:
\begin{align}
g_1^w(r_1^k,t,q) = 	 \begin{cases}
	p_{r_1^k}, & \text{if } ~~ t = t^\text{min+}_{{r_1^k},w} ~~\wedge~~ q = 0 \\ 
	0  		& \text{otherwise}.
	\end{cases} 		\notag	
\end{align}
\begin{align}
g_2^w(r_1^k,t,q) =   \begin{cases}
	1 - p_{r_1^k}, & \text{if } ~~ t = \max( \underline{on}(w), \Gamma_{r_1^k}) ~~\wedge~~ q = 0 \\
	0  & \text{otherwise}.
	\end{cases} 		\notag
\end{align}
The vehicle thus cannot be available for $r_1^k$ at any other location $r' <_R r$: 
\begin{align}
g_1^{r'}(r_1^k,t,q) = 0 	\notag	\\
g_2^{r'}(r_1^k,t,q) = 0	\notag
\end{align}
Concerning $r_1 = \fst(\pi_w)$ the first request of any other waiting location $w \ne \w(r_1^k)$, we use the same trick as for strategy $\mathcal{R}^{q}$ in order to obtain the probabilities for each possible vehicle load $q$:
\begin{align}
g_1^w(r_1,t,q) = \begin{cases}
		p_{r_1} \sum\limits_{t' = \underline{on}(w')}^{ \overline{on}(w')} \Big[ h^{w'}(\prv(r_1),t',q) + \sum\limits_{r' \in \pi_{w'}} h^{r'}(\prv(r_1),t',q)  \Big] , \\
		\hphantom{0}	\hspace{12em}	\text{if } ~~ t = \max( \underline{on}(w),  t^\text{min+}_{{r_1},w} )  \\ 
		0  			\hspace{12em}	\text{otherwise}.
	\end{cases} 	\notag	
\end{align}
\begin{align}
g_2^w(r_1,t,q) = \begin{cases}
		(1 - p_{r_1}) \sum\limits_{t' = \underline{on}(w')}^{ \overline{on}(w')} \Big[ h^{w'}(\prv(r_1),t',q) + \sum\limits_{r' \in \pi_{w'}} h^{r'}(\prv(r_1),t',q)  \Big] , \\
		\hphantom{0}	\hspace{12em} \text{if } ~~ t = \max( \underline{on}(w),  \Gamma_{r_1} )  \\ 
		0  			\hspace{12em} \text{otherwise}.
	\end{cases} 	\notag
\end{align}
with $w' = \w(\prv(r_1))$. 
From any other request $r' <_R r$ we still have:
\begin{align}
g_1^{r'}(r_1,t,q) = 0 	\notag\\
g_2^{r'}(r_1,t,q) = 0 	\notag
\end{align}
For a request $r >_R \fst(\pi_{w}), w \in W^x$:
\begin{align}
g_1^v(r,t, q) = \begin{cases}
		p_{r} \cdot h^v(\prv(r),t, q)   & \text{if } ~~ t >  t^\text{min+}_{{r_1},v}   \\
		p_{r} \cdot \sum_{t' = \underline{on}(w)}^{ t^\text{min+}_{{r_1},v} } h^v(\prv(r),t' , q) ~~~ & \text{if } ~~ t =  t^\text{min+}_{{r_1},v}  \\
		0 & \text{otherwise} . 	
	\end{cases} \notag
\end{align}
\begin{align}
g_2^v(r,t, q) = \begin{cases}
		(1 - p_{r}) \cdot h^v(\prv(r),t, q)   & \text{if } ~~ t > \max( \underline{on}(w),  \Gamma_r )   \\
		(1 - p_{r}) \cdot \sum_{t' = \underline{on}(w)}^{\max( \underline{on}(w),  \Gamma_r ) }    h^v(\prv(r),t' , q) ~~ & \text{if } ~~ t = \max( \underline{on}(w),  \Gamma_r )  \\
		0 & \text{otherwise} . 	
	\end{cases} 	\notag
\end{align}
when replacing $v$ by either $w = \w(r)$ or $r' \in \pi_w, r' <_R r, w = \w(r)$.
\\
\\
At a waiting location $w \in W^x$:
\begin{align}
h^w(r, t, q) ~ = ~  h^w_1(r, t, q) ~ + ~ h^w_2(r, t, q) ~ + ~ h^w_3(r, t, q).	\notag
\end{align}
The aforementioned terms of the sum are:
\begin{align}
h^w_1(r, t, q) = \begin{cases}
		g_1^w(r, t^{w}, q - q_r) \cdot \delta^w(r, t^{w}, q - q_r) \\
			+ ~~ \sum_{\substack{r' \in \pi_w\\r' <_R r}} g_1^{r'}(r, t^{r'}, q - q_r) \cdot \delta^{r'}(r, t^{r'}, q - q_r) , &\text{if } ~~ t - d_{r,w} <  \Gamma_r^\text{next}  \\ 
		0  			&\text{otherwise}.
	\end{cases} 	\notag
\end{align}
where 
\begin{align}
\delta^v(r, t, q) &= \begin{cases}
		1, & \text{if } ~~ t   \le t^\text{max+}_{r,v} ~ \wedge ~ q + q_{r} \le Q \\
		0, & \text{otherwise.}
	\end{cases} 	\notag
\end{align}
and $t^{w}=t - d_{w, r} - s_r - d_{r, w}$, $t^{r'}=t - d_{r', r} - s_r - d_{r, w}$ and $\Gamma_r^\text{next} = \Gamma_r$ if $\nxt(r)$ exists, zero otherwise.
The second term $h^w_2$ is:
\begin{align}
h^w_2(r, t, q) = g_1^w(r, t, q) \cdot \big( 1- \delta(r, t, w) \big) + g_2^w(r, t, q).	 \notag 
\end{align}
Finally:
\begin{align}
h^w_3(r, t, q) =  
		\sum_{\substack{r' \in \pi_w\\r' <_R r}}
		\begin{matrix}
		& \Big[ g_1^{r'}(r, t - d_{r', w}, q) (1 - \delta^{r'}(r,  t - d_{r', w}, q) \big) \\
		& \hspace{3em} + ~~ g_2^{r'}(r, t - d_{r', w}, q) \Big] \cdot  \text{bool}(t - d_{r',w} <  \Gamma_r^\text{next} )
		\end{matrix} \notag
\end{align}
where $\text{bool}(a)$ returns 1 if the Boolean expression $a$ is true, 0 otherwise.
\\
The probability that the vehicle gets rid of request $r$ at $r$'s location is:
\begin{align}
h^r(r, t, q) ~ = ~ 	&g_1^w(r, t - d_{w, r} - s_r, q - q_r) \cdot \delta^w(r, t - d_{w, r} - s_r, q - q_r) \notag \\
			&+  \sum_{\substack{r' \in \pi_w\\r' <_R r}}  g_1^{r'}(r, t - d_{r', r} - s_r, q - q_r) \cdot \delta^{r'}(r, t - d_{r', r} - s_r, q - q_r)	 \notag 
\end{align}
if $t \ge \Gamma_r^\text{next}$, otherwise $h^r(r, t, q) = 0$.
\\
Finally, the probability that request gets discarded from another request $r'$ location:
\begin{align}
h^{r'}(r, t, q) =  \begin{cases}
 			g_1^{r'}(r, t ,q) \cdot (1- \delta^{r'}(r, t, q) ) ~ + ~ g_2^{r'}(r,t,q),	& \text{if } ~~ t \ge \Gamma_r^\text{next} 	\\
 			0, & \text{otherwise.}
	\end{cases} 	\notag
\end{align}

\ACKNOWLEDGMENT{
Computational resources have been provided by the Consortium des Équipements de Calcul Intensif (CÉCI), funded by the Fonds de la Recherche Scientifique de Belgique (F.R.S.-FNRS) under Grant No. 2.5020.11.
Christine Solnon is supported by the LABEX IMU (ANR-10-LABX-0088) of Université de Lyon, within the program "Investissements d’Avenir" (ANR-11-IDEX-0007) operated by the French National Research Agency (ANR).
Finally, we thank Anthony Papavasiliou for its sound suggestions and advices during the early stages of the study.
}

\bibliographystyle{informs2014trsc}
\bibliography{library}

\ECSwitch


\ECHead{E-Companion}

\section{Instance generation\label{sec:instance_generation}}

\subsubsection*{Data used to generate instances}
\begin{figure}[t]
  \begin{minipage}[c]{0.67\textwidth}
    \includegraphics[width=\textwidth]{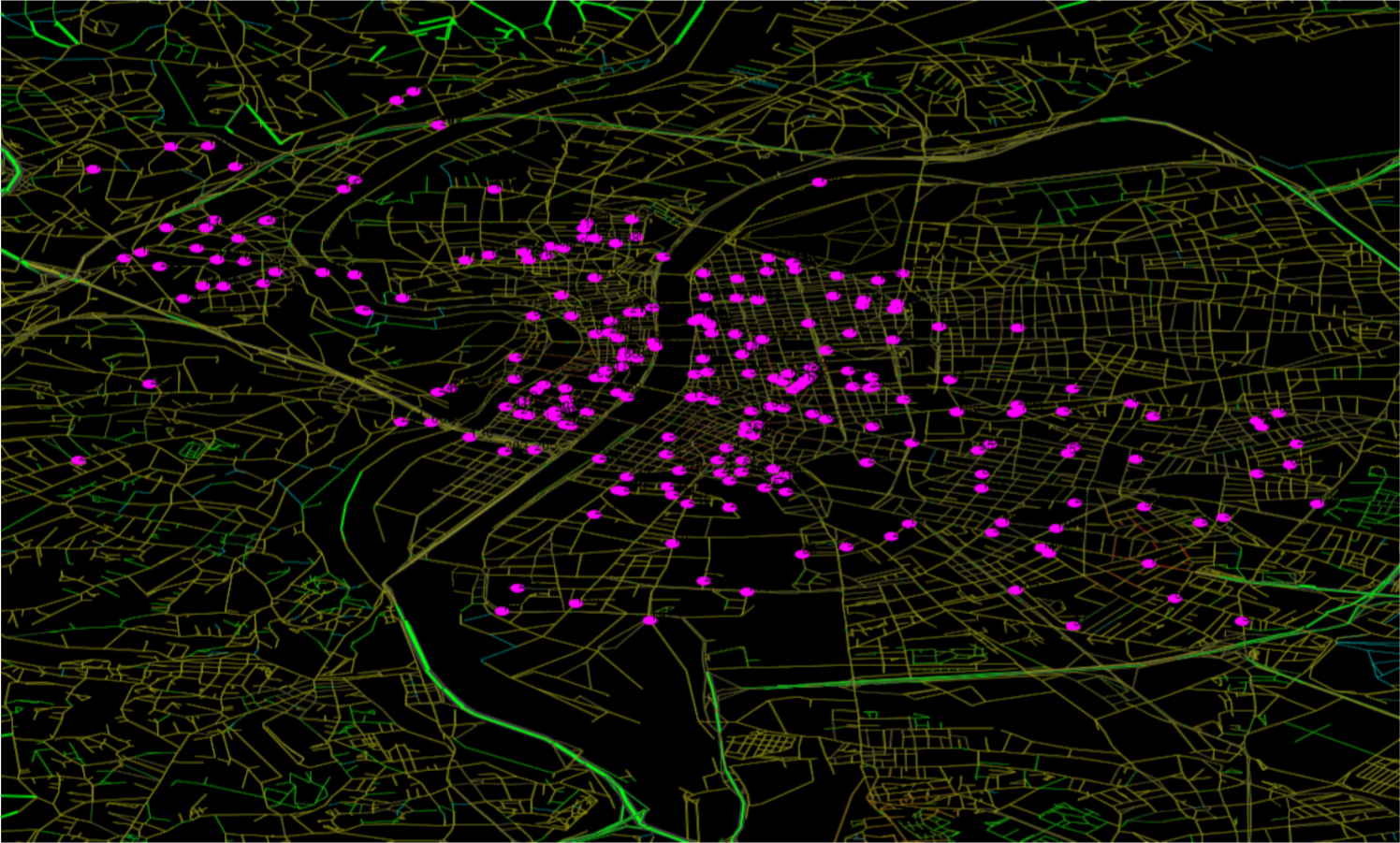}
  \end{minipage}\hfill
  \begin{minipage}[c]{0.3\textwidth}
    \caption{
      Lyon's road network. In purple, the 255 customer vertices.
    } \label{fig:lyon}
  \end{minipage}
\end{figure}

We derive our test instances from the benchmark described in \cite{Melgarejo2015} for the Time-Dependent TSP with Time Windows (TD-TSPTW). This benchmark has been created using real accurate delivery and travel time data coming from the city of Lyon, France.
Travel times have been computed from vehicle speeds that have been measured by 630 sensors over the city (each sensor measures the speed on a road segment every 6 minutes). For road segments without sensors, travel speed has been estimated with respect to speed on the closest road segments of similar type.
Figure \ref{fig:lyon} displays the set of 255 delivery addresses extracted from real delivery data, covering two full months of time-stamped and geo-localized deliveries from three freight carriers operating in Lyon.
For each couple of delivery addresses, travel duration has been computed by searching for a quickest path between the two addresses. In the original benchmark, travel durations are computed for different starting times (by steps of 6 minutes), to take into account the fact that travel durations depend on starting times. In our case, we remove the time-dependent dimension by simply computing average travel times (for all possible starting times). We note $\overline{V}$ the set of 255 delivery addresses, and $d_{i,j}$ the duration for traveling from $i$ to $j$ with $i,j\in \overline{V}$.
This allows us to have realistic travel times between real delivery addresses.
Note that in this real-world context, the resulting travel time matrix is not symmetric. 

\subsubsection*{Instance generation}
We have generated two different kinds of instances: instances with separated waiting locations, and instances without separated waiting locations.
Each instance with separated waiting locations is denoted $n$\textbf{c}-$m$\textbf{w}-$i$, where $n\in\{10,20,50\}$ is the number of customer vertices, $m\in \{5,10,30,50\}$ is the number of waiting vertices, and $x \in [1,15]$ is the random seed.
It is constructed as follows:
\begin{enumerate}
\item We first partition the 255 delivery addresses of $\overline{V}$ in $m$ clusters, using the $k$-means algorithm with $k=m$. During this clustering phase, we have considered symmetric distances, by defining the distance between two points $i$ and $j$ as the minimum duration among $d_{i,j}$ and $d_{j,i}$.
\item For each cluster, we select the median delivery address, {\em i.e.}, the address in the cluster such that its average distance to all other addresses in the cluster is minimal.
The set $W$ of waiting vertices is defined by the set of $m$ median addresses.
\item We randomly and uniformly select the depot and the set $C$ of $n$ customer vertices in the remaining set $\overline{V}\setminus W$.
\end{enumerate}
Each instance without separated waiting locations is denoted $n$\textbf{c}+\textbf{w}-$i$. 
It is constructed by randomly and uniformly selecting the depot and the set $C$ in the entire set $\overline{V}$ and by simply setting $W = C$. 
In other words, in these instances vehicles do not wait at separated waiting vertices, but at customer vertices, and every customer vertex is also a waiting location.

Furthermore, instances sharing the same number of customers $n$ and the same random seed $x$ (\emph{e.g.} 50\textbf{c}-30\textbf{w}-1, 50\textbf{c}-50\textbf{w}-1 and 50\textbf{c+w}-1) always share the exact same set of customer vertices $C$.

\subsubsection*{Operational day, horizon and time slots}
We fix the duration of an operational day to 8 hours in all instances. 
We fix the horizon resolution to $h=480$, which corresponds to one minute time steps.
As it is not realistic to detail request probabilities for each time unit of the horizon (\textit{i.e.}, every minute), we introduce \emph{time slots} of 5 minutes each. 
We thus have $n_{\text{TS}}=96$ time slots over the horizon.
To each \emph{time slot} corresponds a potential request at each customer vertex. 

\subsubsection*{Customer potential requests and attributes.}
\begin{figure}[t]
\centering
\includegraphics[width=1.0\textwidth]{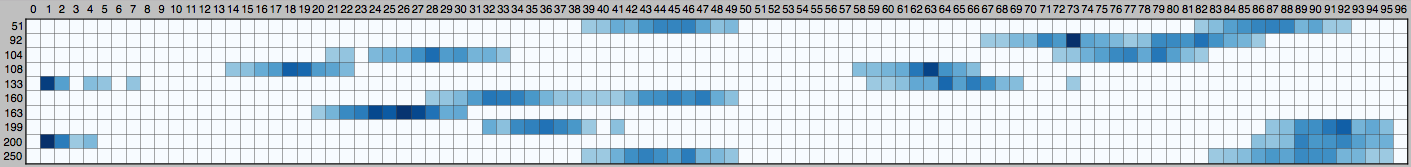}
\caption{
Probability distributions in instance \textit{10-c5w-1}. 
Each cell represents one of the 96 time slots, for each customer vertex.
The darker a cell, the more likely a request to appear at the corresponding time slot.
A white cell represents a zero probability request that is, no potential request.
}
\label{fig:probas_instance}
\end{figure}

For each customer vertex $c$, we generate the request probabilities associated with $c$ as follows.
First, we randomly and uniformly select two integer values $\mu_1$ and $\mu_2$ in $[1,n_{\text{TS}}]$. 
Then, we randomly generate 200 integer values: 100 with respect to a normal distribution the mean of which is $\mu_1$ and 100 with respect to a normal distribution the mean of which is $\mu_2$. 
Let us note $nb[i]$ the number of times value $i\in [1,n_{\text{TS}}]$ has been generated among the 200 trials. 
Finally, for each reveal time $\Gamma\in H$, if $\Gamma\mod 5 \neq 0$, then we set $p_{(c,\Gamma)}=0$ (as we assume that requests are revealed every 5 minute time slots). 
Otherwise, we set $p_{(c,\Gamma)} = \min(1,\frac{nb[\Gamma/5]}{100})$.
Hence, the expected number of requests at each customer vertex is smaller than or equal to 2 (in particular, it is smaller than 2 when some of the 200 randomly generated numbers do not belong to the interval $[1,n_{\text{TS}}]$, which may occur when $\mu_1$ or $\mu_2$ are close to the boundary values).
Figure \ref{fig:probas_instance} shows a representation of the distributions in an instance involving 10 customer vertices. 

For a same customer vertex, there may be several requests on the same day at different time slots, and their probabilities are assumed independent.
To each potential request $r=(c_r,\Gamma_r)$ is assigned a deterministic demand $q_r$ taken uniformly in $[0,2]$, a deterministic service duration $s_r=5$ and a time window $[\Gamma_r, \Gamma_r+\Delta-1]$, where $\Delta$ is taken uniformly in $\{5,10,15,20\}$ that is, either 5, 10, 15 or 20 minutes to meet the request. Note that the beginning of the time window of a request $r$ is equal to its reveal time $\Gamma_r$.
This aims at simulating operational contexts similar to the practical application example described in section \ref{sec:Introduction} (the on-demand health care service at home), requiring immediate responses within small time windows.

\end{appendices}

\end{document}